\documentclass{article}
\usepackage{amsmath}
\usepackage{subfig}
\usepackage{lipsum}
\usepackage{algpseudocode}
\usepackage{placeins}
\usepackage{blindtext}
\usepackage{booktabs}
\usepackage{tcolorbox}
\usepackage{stfloats}
\algnotext{EndFor}
\algnotext{EndIf}

\DeclareMathOperator*{\minimize}{minimize}
\usepackage{hyperref}

\usepackage[accepted]{icml2023}

\usepackage{amsmath,amsfonts,bm}

\def\1{\bm{1}}

\DeclareMathAlphabet{\mathsfit}{\encodingdefault}{\sfdefault}{m}{sl}
\SetMathAlphabet{\mathsfit}{bold}{\encodingdefault}{\sfdefault}{bx}{n}

\input{mysymbol.sty}

\usepackage[utf8]{inputenc} 
\usepackage[T1]{fontenc}    
\usepackage{hyperref}       
\usepackage{url}            
\usepackage{booktabs}       
\usepackage{amsfonts}       
\usepackage{nicefrac}       
\usepackage{microtype}      
\usepackage{xcolor}         

\begin{document}
\twocolumn[
\icmltitle{Automatic Data Augmentation via Invariance-Constrained Learning}



\begin{icmlauthorlist}
\icmlauthor{Ignacio Hounie}{upenn}
\icmlauthor{Luiz F. O. Chamon}{stutgart}
\icmlauthor{Alejandro Ribeiro}{upenn}
\end{icmlauthorlist}
\icmlaffiliation{upenn}{University of Pennsylvania}
\icmlaffiliation{stutgart}{University of Stuttgart}
\icmlcorrespondingauthor{Ignacio Hounie}{ihounie@seas.upenn.edu}

\icmlkeywords{Machine Learning, ICML, Data Augmentation, Constrained Learning, Invariance}
\vskip 0.3in
]
\icmltitlerunning{Data Augmentation via Invariance-Constrained Learning}
\printAffiliationsAndNotice{}
\begin{abstract}
Underlying data structures, such as symmetries or invariance to transformations, are often exploited to improve the solution of learning tasks. However, embedding these properties in models or learning algorithms can be challenging and computationally intensive. Data augmentation, on the other hand, induces these symmetries during training by applying multiple transformations to the input data. Despite its ubiquity, its effectiveness depends on the choices of which transformations to apply, when to do so, and how often. In fact, there is both empirical and theoretical evidence that the indiscriminate use of data augmentation can introduce biases that outweigh its benefits. This work tackles these issues by automatically adapting the data augmentation  while solving the learning task. To do so, it formulates data augmentation as an invariance constrained learning problem and leverages Monte Carlo Markov Chain (MCMC) sampling to solve it. The result is an algorithm that not only does away with a priori searches for augmentation distributions, but also dynamically controls if and when data augmentation is applied. We validate empirically our theoretical developments in automatic data augmentation benchmarks for CIFAR and ImageNet-100 datasets. Furthermore, our experiments show how this approach can be used to gather insights on the actual symmetries underlying a learning task.
\end{abstract}
\section{Introduction}
Exploiting the underlying structure of data is a key principle of data analysis. Its use has been fundamental to the success of machine learning solutions, from the translational equivariance of convolutional neural networks~\citep{fukushima} to the invariant attention mechanism in Alphafold~\citep{alphafold}. However, embedding invariances and symmetries in model architectures is hard in general and when possible, often incurs a high computational cost. This is the case, of rotation invariant neural network architectures that rely on group convolutions, which are feasible only for small, discrete transformation spaces or require coarse undersampling due to their high computational complexity~\citep{Taco-OG, finzi-lie}.

A widely used alternative consists of modifying the data rather than the model. That is, to~\emph{augment} the dataset by applying transformations to samples in order to induce the desired symmetries or invariances during training. Data augmentation, as it is commonly known, is used to train virtually all state-of-the-art models in a variety of domains~\citep{data-augmentation-survey}. This empirical success is supported by theoretical results showing that, when the underlying data distribution is invariant to the applied transformations, data augmentation provides a better estimation of the statistical risk~\citep{Dobri-data-aug,group-invariance-bounds, Gal-benefits-of-invariance, PAC-invariance-blum}. On the other hand, applying the wrong transformations can introduce biases that may outweigh these benefits~\citep{Dobri-data-aug, PAC-invariance-blum}. 

Choosing which transformations to apply, when to do so, and how often, is thus paramount to achieving good results. However, it requires knowledge about the underlying distribution of the data that is typically unavailable in learning settings. Several approaches to learning an augmentation policy or distribution over a given set of transformations exist, such as reinforcement learning~\citep{Auto-augment}, genetic algorithms~\citep{population-based-augmentation}, density matching~\citep{fast-auto-augment, rand-augment, faster-auto-augment}, gradient matching~\citep{deep-aa}, bi-level optimization~\citep{bi-level-dada-augmentation, Differentiable-aug-direct}, jointly optimizing over transformations using regularised objectives~\citep{augerino-finzi}, variational bayesian inference~\citep{trm-vagelis}, bayesian model selection~\citep{invariance_laplace} and alignment regularization~\citep{representation-alignment}. Nevertheless, optimization based methods often require computing gradients with respect to transformations~\citep{trm-vagelis, bi-level-dada-augmentation} and several of these methods resort to computationally intensive search phases, optimization of auxiliary models, or additional data, while failing to outperform fixed user-defined augmentation distributions~\citep{trivial-augment}.

 In this work, we formulate data augmentation as an invariance-constrained learning problem. That is, we specify a set of transformations and a desired level of invariance or robustness with respect to these transformations, and recover an augmentation distribution that imposes this requirement on the learned model. We do this in a non-parametric fashion, i.e., without explicitly parametrising the distribution over transformations.
 
 More specifically, our invariance requirement is weighted by the probability of each data point. Hence, we require the output of our model to be stable only on the support of the underlying data distribution, and more so on common samples. By imposing this requirement as a constraint on the learning task and leveraging recent duality results, the amount of data augmentation can be automatically adjusted during training. An advantage of the constrained learning formulation is that it mitigates the potential biases introduced by data augmentation without doing away with its potential benefits. We propose an algorithm that combines stochastic primal-dual methods and MCMC sampling to do away with the need for transformations to be differentiable. We evaluate our method in automatic data augmentation benchmarks in CIFAR and ImageNet-100 datasets. Furthermore, we show how our method provides insights on the actual symmetries underlying a learning task, using synthetically invariant MNIST and Fashion-MNIST datasets.
 
 %

\section{Data Augmentation in Supervised Learning}

As in the standard supervised learning setting, let $\mathbf{x} \in \mathcal{X} \subseteq \mathbb{R}^{d}$ denote a feature vector and $y \in \mathcal{Y} \subseteq \mathbb{R}$ its associated label or measurement. For classification tasks, we take $\mathcal{Y} \subseteq \mathbb{N}$. Let~$\mathfrak{D}$ denote a probability distribution over the data pairs~$(\mathbf{x}, y)$ and~$\ell: \mathcal{Y}\times \mathcal{Y} \to  \mathbb{R}_+$ be a non-negative, convex loss function, e.g., the cross entropy loss. Our goal is to learn a predictor $f_{\boldsymbol{\theta}}: \mathcal{X} \to \mathcal{Y}$ in some hypothesis class $\mathcal{H}_{\boldsymbol{\theta}}=\left\{f_{\boldsymbol{\theta}} \mid \boldsymbol{\theta} \in \Theta \subseteq \mathbb{R}^{p} \right\}$ that minimizes the expected loss, 
\begin{align}\tag{SRM}\label{SRM}
\minimize_{\boldsymbol{\theta} \in \Theta} \; R(f_{\boldsymbol{\theta}}):=\mathbb{E}_{(\mathbf{x}, y) \sim \mathfrak{D}}[\ell(f_{\boldsymbol{\theta}}(\mathbf{x}), y)].
\end{align}
We consider the distribution $\mathfrak{D}$ to be unknown, except for the dataset $\{(\mathbf{x}_i, y_i), i= 1,\ldots,n\}$ of $n$ i.i.d. samples from $\mathfrak{D}$. Therefore, we rely on the empirical approximation of the objective of (SRM), 
\begin{align}\label{ERM}
\hat{R}(f_{\boldsymbol{\theta}}):=\frac{1}{N}\sum_{i=1}^n \ell(f_{\boldsymbol{\theta}}(\mathbf{x}_i), y_i). 
\end{align}
One of the aims of data augmentation is to improve the approximation $\hat{R}$ of the statistical risk $R$ when dealing with a dataset that is not sufficiently representative of the data distribution. To do so, we consider transformations of the feature vector $g: \mathcal{X}\to\mathcal{X}$, taken from the (possibly infinite) transformation set $\mathcal{G}$. Common examples include rotations and translations in images. Data augmentation leverages these transformations to generate new data pairs $(g\mathbf{x}, y)$ by sampling transformations according to a probability distribution $\mathfrak{G}$  over $\mathcal{G}$, leading to the learning problem
\begin{align}\label{Aug-ERM}
\minimize_{\boldsymbol{\theta} \in \Theta} \; \hat{R}_{\text{aug}}(f_{\boldsymbol{\theta}}):=\frac{1}{N}\sum_{i=1}^N  \mathbb{E}_{g \sim \mathfrak{G}}\left[\;\ell(f_{\boldsymbol{\theta}}(g\mathbf{x}_i), y_i)\right].
\end{align}

Note that the empirical risk approximation $\hat{R}$ in (\ref{ERM}) can be interpreted as an approximation of the data distribution $\mathfrak{D}$ by a discrete distribution that places atoms on each data point. In that sense, $\hat{R}_\text{aug}$ in~(\ref{Aug-ERM}) can be thought of as the Vicinal Risk Minimization~\citep{vicinal-risk-minimization} counterpart of~(\ref{ERM}), in which the atoms on $\mathbf{x}_i$ are replaced by a local distribution over the transformed samples $g\mathbf{x}_i$, i.e.,
\begin{align}\label{VRM}
\hat{R}_{\text{aug}}(f_{\boldsymbol{\theta}}) = \frac{1}{N}\sum_{i=1}^N \int \ell\left(f_{\boldsymbol{\theta}}(g\mathbf{x}_i), y_i\right) d P(g\mathbf{x}_i),
\end{align}
where the measure $P$ over $\mathcal{X}$ is induced by the distribution $\mathfrak{G}$ over $\mathcal{G}$. As it can be seen from~(\ref{VRM}) if $\mathfrak{G}$ is not chosen adequately, $\hat{R}_{aug}$ can be a poor estimate of $R$, introducing biases that outweigh the benefits of data augmentation~\citep{Dobri-data-aug,PAC-invariance-blum}. On the other hand, if the data distribution $\mathfrak{D}$ is statistically invariant under the action of $\mathcal{G}$, and $\mathfrak{G}$ is chosen so as to induce those invariances in the solution, then learning using (\ref{Aug-ERM}-\ref{VRM}) has provable advantages in terms of sample complexity~\citep{Dobri-data-aug,bietti-group-invariance, group-invariance-bounds, Gal-benefits-of-invariance}.

In this work, we tackle the choice of $\mathfrak{G}$ given a set of transformations $\mathcal{G}$, i.e., how to sample transformations so that the solution of the learning problem is approximately invariant, as defined on the next section. Unlike \emph{invariance learning}~\citep{Convex-Invariance-Learning, finn-metalearning-invariance, augerino-finzi, invariance_laplace}, we do not seek to learn the transformations $\mathcal{G}$ from the data. We also explore how to incorporate the expectation of the risk of transformed samples in the learning algorithm, so that even when the distribution is not invariant --- or the hypothesis space is not rich enough to capture invariances of the data --- we can still avoid introducing a bias. Note that invariance to transformations in $\mathcal{G}$ may hold for the true distribution or may also be a desirable property of the solution, for example, to achieve robustness~\citep{Geometric-robustness, generalizing-adv-data-aug, semantic-adversarial}. 
%

\section{From Invariance to an Augmentation Distribution}
%

It can be straightforward to specify a set of transformations~$\ccalG$ to which the solution should be~\emph{approximately} invariant~(e.g., image rotations and translations). However, finding a transformation distribution~$\mathfrak{G}$ that leads to the desired properties in the solution can be challenging. What is more, using a fixed~$\mathfrak{G}$ as in~(\ref{Aug-ERM}) prevents us from controlling when and how much augmentation is used during training, running the risk of biasing the final solution. 

In the next sections, we explain how a data augmentation distribution can be obtained from imposing invariance. We first argue how \emph{invariance} leads to a robustness requirement~(Section~\ref{proposed:invariance-as-robustness}) and then show that this invariance constrained learning problem yields an augmentation distribution~(Section~\ref{proposed:robustness-to-aug}). 


\subsection{From Invariance to Robustness}\label{proposed:invariance-as-robustness}

If the data is invariant to action $g$, we want to learn a model that respects this invariance. This induces the requirement to learn a model such that $f_{\bbtheta}(\bbx) = f_{\bbtheta}(g\bbx)$ for all~$g \in \ccalG$. Here, we focus on learning models in which invariance requirements are stated in terms of the loss function and we therefore require that the model satisfies
\begin{align}\label{eqn_loss_invariance}
\ell(f_{\bbtheta}\left(\bbx\right), y) = \ell(f_{\bbtheta}\left(g\bbx\right), y), \; \forall \; g \in \ccalG \; .
\end{align}
If the model is indeed invariant in the sense that $f_{\bbtheta}(\bbx) = f_{\bbtheta}(g\bbx)$, \eqref{eqn_loss_invariance} holds as well. The advantage of \eqref{eqn_loss_invariance} is that it explicitly incorporates the structure of the learning task. This is useful when the model is not perfectly invariant, in which case the difference $\ell(f_{\bbtheta}(g\bbx), y) -\ell(f_{\bbtheta}(\bbx), y)$ measures the cost of not having invariance in terms of the loss function. We leverage this observation to define the invariance loss as 
\begin{align}\label{maxg}
\ell_{\text{inv}}(f_{\bbtheta}, \bbx, y) :=\max_{g \in \ccalG} \ell(f_{\bbtheta}\left(g\bbx\right), y) - \ell(f_{\bbtheta}\left(\bbx\right), y).
\end{align}
This loss is always nonnegative because the identity action is assumed to be part of the invariant set $\ccalG$. In fact, the set $\ccalG$ defines an equivalence class of points $g\bbx$ that we know map to the same output $y$. The loss $\ell_{\text{inv}}$ defined in \eqref{maxg} is a measurement of how far the learnable model $f_{\bbtheta}$ is from defining a corresponding equivalence class with respect to the loss. We can therefore define an invariance risk by taking the average of the invariance loss in \eqref{maxg} with respect to the data distribution,
\begin{align}~\label{worst-case}
   R_\text{inv}(f_{\bbtheta}) 
      :=  \mbE_{(\bbx, y) \sim \mathfrak{D}}
             \left[\ell_{\text{inv}}(f_{\bbtheta}, \bbx, y) \right].
\end{align}
It is interesting to substitute \eqref{maxg} into \eqref{worst-case}. Using the linearity of expectations we can conclude that the invariance risk is
\begin{align}\label{r-inv-bound}
   R_\text{inv}\left( f_{\bbtheta}\right)
      ~=~ &   \mbE_{(\bbx, y) \sim \mathfrak{D}}
               \left[ \max_{g \in \ccalG} \ell(f_{\bbtheta}\left(g\bbx\right), y) \right] \nonumber \\
        & \qquad
          - \mbE_{(\bbx, y) \sim \mathfrak{D}}
               \left[\ell(f_{\bbtheta}\left(\bbx\right), y)\right].
\end{align}
The second term in \eqref{r-inv-bound} is the standard statistical risk without data augmentation. The first term can be interpreted as a data augmentation distribution [cf. \eqref{Aug-ERM}] in which all of the augmentation mass is allocated to the element $g\bbx$ of the invariance set $\ccalG$ for which the loss is highest. This term can also be interpreted as a robust adversarial loss, e.g.,~\cite{madry-adversarial}, in which the adversary chooses the element $g\bbx\in\ccalG$ with the highest loss.

Note that minimizing the invariance risk $R_\text{inv}$ does not necessarily lead to a good model $f_{\bbtheta}$ but rather one that has constant loss under transformations of the input. This motivates a problem formulation in which we want to find models $f_{\bbtheta}$ with small statistical risk $R$ [cf. \eqref{SRM}] and small invariance risk  $R_\text{inv}$ [cf. \eqref{r-inv-bound}]. We therefore combine these two terms to propose the constrained learning problem
\begin{align}\tag{CSRM'}\label{CSRM-og}
P^{\star} \;=\; &\min_{\bbtheta \in \Theta} \quad \mbE_{(\bbx, y) \sim \mathfrak{D}}\left[\ell(f_{\bbtheta}(\bbx), y)\right]\\
& \text{s. to } \quad \mbE_{(\bbx, y) \sim \mathfrak{D}}\left[ \max_{g \in \ccalG}\left[ \ell(f_{\bbtheta}(g\bbx), y) \right] \right] \leq \epsilon \;.\nonumber
\end{align}
Observe that in the constraint in \eqref{CSRM-og} we keep the first term in \eqref{r-inv-bound} only. We do that because to reduce \eqref{r-inv-bound} in a problem in which the standard risk is being minimized, it suffices to reduce the first term of \eqref{r-inv-bound}.

The formulation in \eqref{CSRM-og} tackles the two terms forming the invariant risk bound in~(\ref{r-inv-bound}), but instead of combining them directly, it incorporates the data augmentation term as a constraint in the typical statistical risk minimization problem~(\ref{SRM}). This formulation has the advantage that if a solution to the unconstrained problem is feasible, i.e., satisfies the invariance constraint in~(\ref{CSRM-og}), the presence of that constraint has no effect on the statistical problem. Still, it can be beneficial when approximating the solution of~(\ref{CSRM-og}) empirically. We will explore this fact in the next section, where we tackle the practical challenges involved in solving~(\ref{CSRM-og}).

\subsection{From Robustness to an Augmentation Distribution}\label{proposed:robustness-to-aug}

Solving the maximization in the constraint can be difficult when $\ccalG$ is not finite or $f_{\bbtheta}$ is a deep neural network. Even when the transformation space is low dimensional, as is the case of translations and rotations, the highly non-convex loss landscape of these models makes the maximization over $\ccalG$ challenging~\citep{madry-spatial-robustness}. As shown by~\cite{Semi-Inf}, the maximisation of the loss over transformations can be written as the semi-infinite constrained optimization problem
 \begin{align}
\max_{g \in \ccalG}\ell\left(f_{\bbtheta}\left(g\bbx\right), y\right)  = & \sup_{\lambda \in \mathcal{L}^2_+} \int_{\ccalG} \lambda(g)\ell(f_{\bbtheta}(g\bbx), y)dg\\ \nonumber
 & \text{s. to}  \int_{\ccalG} \lambda(g)dg = 1.\;\;
 \end{align}
Notice that the solution of this optimization problem~$\lambda^{\star}(g)$ is a non-negative, normalized function and  can therefore be interpreted as a distribution over transformations that depends on the sample point~$(\bbx, y)$ as well as the model~$f_{\bbtheta}$. This allows us to re-interpret the maximization over $\ccalG$ as an expectation, i.e.,
 \begin{align}\label{lambda_star}
&\mbE_{(\bbx, y) \sim \mathfrak{D}}\left[ \max_{g \in \ccalG} \ell(f_{\bbtheta}\left(g\bbx\right), y) \right] 
\\ & \qquad\qquad\qquad
= \mbE_{(\bbx, y) \sim \mathfrak{D}}\left[\mbE_{g \sim \lambda^{\star}}\left[ \ell(f_{\bbtheta}(g\bbx), y) \right]\right]. &\nonumber
 \end{align}
 
 Note that this risk upper bounds the risk under any augmentation distribution defined on $\ccalG$.
 
 Then we can re-write problem~(\ref{CSRM-og}) replacing its robustness constraint with one using data augmentation according to $\lambda^{\star}$, namely,
\begin{tcolorbox}
\vspace{-0.15in}
\begin{align}\tag{CSRM}\label{CSRM}
P^{\star} \;=\; &\min_{\bbtheta \in \Theta} \quad \mbE_{(\bbx, y) \sim \mathfrak{D}}\left[\ell(f_{\bbtheta}(\bbx), y)\right]\\
& \text{s. to } \quad \mbE_{(\bbx, y) \sim \mathfrak{D}}\left[\mbE_{g \sim \lambda^{\star}}\left[ \ell(f_{\bbtheta}(g\bbx), y) \right] \right] \leq \epsilon \;.\nonumber
\end{align}
\end{tcolorbox}
For conciseness in~(\ref{CSRM}) we have included only one invariance constraint associated with a single set of transformations $\ccalG$. However, our approach can be extended to an arbitrary number of constraints defined by transformation sets $\ccalG_i, \; i=1,\ldots,m$ (each inducing an augmentation distribution $\lambda^{\star}_i$), and constraint levels $\epsilon_i$. All of the following derivations still hold, regardless of the number of constraints.

\section{Algorithm Development}

Solving~(\ref{CSRM}) presents two challenges. First, it is a constrained statistical learning problem, which involves the unknown data distribution $\mathfrak{D}$. We address this by resorting to an empirical dual problem as explained in Section~\ref{proposed:primal-dual}.
 Second, it can be hard to sample from $\lambda^{\star}$.  We address this by introducing a smooth approximation that leverages MCMC methods in Section~\ref{proposed:smoothed-invariance}.

\subsection{Dual Empirical Constrained Learning}\label{proposed:primal-dual}

To tackle the invariance-constrained statistical risk minimization problem, we leverage recent duality results in constrained learning theory~\citep{Chamon-PAC, non-convex}, that approximate (\ref{CSRM}) by its empirical dual, 
\begin{align}
D^{\star}_{\text{emp}} = \max_{\gamma \geq 0 } \min_{\bbtheta \in \Theta} \frac{1}{n}\sum_{i=1}^n \big[ & \ell(f_{\bbtheta}(\bbx_i), y_i) \nonumber \\ 
& + \gamma \left(\mbE_{g \sim \lambda^{\star}}\left[ \ell(f_{\bbtheta}(g\bbx_i), y_i) \right]- \epsilon \right) \big].\nonumber
& \\ \tag{DE-CRM}~\label{D-CERM}
\end{align}
The advantage of~(\ref{D-CERM}) is that it is an unconstrained problem that, provided we have enough samples and the parametrization is rich enough, can approximate the constrained statistical problem (\ref{CSRM}). Namely, the difference between the optimal value of the empirical dual $D^{\star}_{\text{emp}}$ and the statistical primal $P^{\star}$, i.e., the empirical duality gap is bounded~\citep{non-convex} with high probability.

As in regular data augmentation, we will also approximate the expectation over $\lambda^{\star}$ by sampling transformations as discussed in Section~\ref{proposed:smoothed-invariance}. Then, the problem~(\ref{D-CERM}) becomes an unconstrained deterministic problem that can be solved using the algorithm described in Section~\ref{proposed:algorithm}.

Note that finding a Lagrangian minimizer for a fixed value of the dual variable ($\gamma$) is equivalent to minimising the risk under a fixed mixture augmentation distribution
or a penalised or regularised learning objective. However, solving the constrained problem, namely maximising over $\gamma$, has fundamental differences. 

First, constraints explicit the requirement they represent. While the degree of invariance imposed should depend only on the statistical problem at hand, the value of $\gamma$ needed to achieve it will depend on the sample size, the parametrization and the learning algorithm. In contrast, constrained learning dynamically adjusts the amount of augmentation --- dictated by $\gamma$ --- to a particular learning setup. Second, the optimal dual variable can give information about the trade-off minimising the loss over training samples and satisfying the invariance constraint. In penalised approaches, on the contrary, this trade-off is fixed. Lastly, the aforementioned informativeness and interpretability can facilitate hyper-parameter tuning. The insights gathered from optimal dual variables can be leveraged a posteriori, for instance, to manually choose appropriate transformations, relax the invariance constraint levels, or change the learning setup (e.g., increase the capacity of the model class). 

\subsection{Sampling towards invariance}\label{proposed:smoothed-invariance}

 If the optimal distribution $\lambda^{\star}$ is not smooth, it is challenging to sample from it with sufficient accuracy~\citep{sampling-review-stochastic-optim}. Consequently, obtaining an unbiased estimator of $\mbE_{g \sim \lambda^{\star}}\left[ \ell(f_{\bbtheta}(g\bbx), y) \right]$ may not be possible. Therefore, we add an $L_2$ norm penalisation, which promotes smoothness, to leverage MCMC methods.

We then define the \emph{c-smoothed distribution} $\lambda^{\star}_c$ as a solution to the regularised problem
 \begin{align*}
&\lambda^{\star}_c \subseteq \\
& \;\; \operatorname{argmax}_{\lambda \in \mathcal{L}^2_+} \int_\ccalG \lambda(g)\ell(f_{\bbtheta}(g\bbx), y)  dg + c\int_\ccalG \lambda(g)^2 dg, \\ \nonumber
 &\;\; \; \text{   s. to} \;   \int_\ccalG \lambda(g)dg = 1\;\;
  \normalsize
 \end{align*}
The regularization term introduces an optimality gap with respect to worst case perturbations, i.e.,~$\mbE_{g \sim \lambda^{\star}_c}\left[ \ell(f_{\bbtheta}(g\bbx), y) \right] \leq \max_{g \in \ccalG}\ell\left(f_{\bbtheta}\left(g\bbx\right), y\right)$. However, for particular values of $c$ the regularized problem has a closed form solution~\citep{Semi-Inf} that allows us to sample from it easily. Namely, there exists a constant $c \geq 0$ such that $\lambda^{\star}_c(\bbx, y, g) \propto \ell(f_{\bbtheta}(g\bbx), y)$.

We need not find its partition function to sample from $\lambda^{\star}_c$, and can instead leverage Monte Carlo Markov Chain methods (MCMC). MCMC methods~\citep{MCMC-og} are based on constructing a Markov chain that has the target distribution as an equilibrium distribution. \emph{Independent} Metropolis Hastings uses a state independent --- usually fixed --- proposal for each step. In our case, it only requires applying a transformation and computing a forward pass of the neural network to evaluate the loss. This enables the use of non-differentiable transformations, and has the advantage that the density at consecutive proposals can be evaluated in parallel, allowing speedups in the sampling step.

MH methods thus allow to sample the proposal distribution with low computational cost. It also allows the computation of transformations on CPU, unlike gradient based methods, thus requiring minimal modifications to typical training pipelines. Although MH methods exhibit random walk behaviour, which leads to slow convergence in high dimensional settings~\citep{MCMC-intro, adaptive-ind-mh}, the space of augmentation transformations is typically low dimensional. For example, $d=18$ for the experiments in Section~\ref{experiments}. Furthermore, even coarse sampling approximations can yield useful augmentation distributions in practice (see Section~\ref{experiments}). 

However, nothing precludes our method from being used with other sampling algorithms. For example, Hamiltonian Monte Carlo methods (see e.g. ~\cite{HMC}), namely Langevin Monte Carlo~\cite{langevin-OG, Langevin} samplers, are suitable when using differentiable transformations.

Then, we can obtain a set of $m$ samples drawn from $\lambda^{\star}_c$ and approximate the expectation over the group by the sample mean
\begin{align*}
   \mbE_{g \sim \lambda^{\star}_c}\left[ \ell(f_{\bbtheta}(g\bbx_i), y_i) \right] \approx  \frac{1}{m}\sum_{j=1}^m \ell(f_{\bbtheta}(\mathbf{g_{j} \bbx_i}), y_i),
\end{align*}
where $g_{1}, \ldots, g_{m} \overset{\text{i.i.d.}}{\sim}\ell(f_{\bbtheta}(g\bbx_i), y_i)$ are $m$ transformations sampled from the smoothed distribution $\lambda^{\star}_c(f_{\bbtheta}, \bbx_i, y_i)$.

Algorithm~\ref{algo-mh} describes the implementation of independent-MH with a uniform proposal. By keeping only one sample ($m = 1$) we recover the usual augmentation setting, that yields one augmentation per sample in the training batch. In our experiments we address this setting, because it is extensively used in practice. However, simply keeping more samples from the chain ($m>1$) allows to extend the method to the batch augmentation setting~\citep{batch-augmentation}, which creates several augmented samples from the same instance in each batch.

 \begin{algorithm}
\caption{Independent MH sampler}\label{algo-mh}
\begin{algorithmic}[1] 
\State $g^{(0)}\sim \mathcal{U}(\ccalG)$ \Comment{Sample initial state}
\State  $\ell^{(0)} = \ell\left(f_{\bbtheta}\left(g^{(0)}\bbx\right), y\right)$ \Comment{Evaluate loss}
\For{$t = 1, \ldots, n_{steps}$}
\State $ g_{\text{prop}}\sim \mathcal{U}(\ccalG)$ \Comment{Sample next proposal}
\State  $\ell_{\operatorname{prop}} = \ell\left(f_{\theta}\left(g_{\text{prop}}\bbx\right), y \right)$ \Comment{Evaluate Loss}
\State $p = \min \left(1, \frac{\ell_{\operatorname{prop}}}{\ell^{(t-1)}}\right) $ \Comment{Acceptance Prob}
\State w.p. $p$:\Comment{Accept/Reject}
\State$\quad g^{(t)} = g_{\text{prop}}$, $\ell^{(t)} = \ell_{\operatorname{prop}} $
\State else:
\State$\quad g^{(t)} = g^{(t-1)}\;$, $\ell^{(t)} = \ell^{(t-1)}$
\EndFor
\end{algorithmic}
\end{algorithm}

\subsection{Primal-Dual Algorithm}\label{proposed:algorithm}

Since the cost of the inner minimization, i.e. training the model, can be high, we adopt an alternating update scheme \citep{Uzawa-alternate} for the primal and dual variables, as in  \citep{non-convex, Fioretto2020LagrangianDF}. 

A bounded empirical duality gap \emph{does not} guarantee that the primal variables obtained after running the alternating primal-dual Algorithm~\ref{algo-pd} and solving the saddle point problem approximately are near optimal or approximately feasible. Although stronger \emph{primal recovery} guarantees can be obtained by randomizing the learning algorithm~\citep{non-convex}, it requires storing model parameters $\bbtheta$ at each iteration and there is empirical evidence \citep{non-convex,Semi-Inf, Elenter, CLIMB, stable-gnns, domain-model-2} that good solutions can still be obtained without randomization.


\begin{algorithm}
\caption{Primal-Dual Augmentation}\label{algo-pd}
\begin{algorithmic}[1] 
\State $ \lambda = 0, \; \bbtheta =\bbtheta_0 $.
\For{$\operatorname{Batch} \text{ in }\left\{\left(\bbx_{i}, y_{i}\right)\right\}_{i=1}^{n}$}
\For{$\left(\bbx_{i}, y_{i}\right) \;  \in \;$Batch}
\Comment{Sample transformations}
\State $ g_{i1}, \ldots, g_{im} \sim^{iid} \ell(f_{\bbtheta}(g\bbx_i), y_i)$
\EndFor
\State 
$\ell_\text{c}=\frac{1}{|\text{Batch}|}\sum\limits_{\left(\bbx_{i}, y_{i}\right) \;  \in \; \text{Batch}} \left[\frac{1}{m}\sum_{j=1}^m \ell(f_{\bbtheta}(g_{ij} \bbx_i), y_i) \right]$
\State 
$s= \ell_\text{c} - \epsilon$\Comment{Slack}

\State 
$\ell=\frac{1}{|\text{Batch}|} \sum_{\left(\bbx_{i}, y_{i}\right) \;  \in \; \text{Batch}}  \ell(f_{\bbtheta}(\bbx_i), y_i)$
\Comment{Loss}

\State
$
\hat{L} = \ell + \gamma s
$\Comment{Lagrangian}
\State
$\bbtheta = \bbtheta-\eta_{p} \nabla_{\bbtheta} \hat{L}$ \Comment{Primal update}

\State
$\gamma=\left[\gamma +\eta_d s\right]_{+}$ \Comment{Dual Update}

\EndFor
\end{algorithmic}
\end{algorithm}

\begin{table*}[t!]
\centering
\setlength{\tabcolsep}{5pt} 
\renewcommand{\arraystretch}{1.0}
\scalebox{0.9}{\begin{tabular}{l|ccc|ccc}
\toprule
& \multicolumn{3}{c|}{Standard} & \multicolumn{3}{c}{Wide}\\
&TA &  DeepAA & \textbf{OURS} & TA &  DeepAA & \textbf{OURS} \\ 
\hline CIFAR10 & &  &  &  & &\\ 
Wide-ResNet-40-2&$96.55 \pm 0.11$&$96.43 \pm 0.09$&$\mathbf{96.76 \pm 0.14}$&$96.24 \pm 0.19$&$96.27 \pm 0.19$&$\mathbf{97.05 \pm 0.18}$ \\ 
Wide-ResNet-28-10&$97.46 \pm 0.10$&$97.57 \pm 0.15$&$\mathbf{97.74 \pm 0.10}$&$97.51 \pm 0.20$&$97.27 \pm 0.10$&$\mathbf{97.85 \pm 0.17}$ \\ 
\hline CIFAR100 & &  &  &  & & \\ 
Wide-ResNet-40-2&$79.92 \pm 0.13$&$79.45 \pm 0.42$&$\mathbf{80.83 \pm 0.31}$&$79.96 \pm 0.45$&$79.36 \pm 0.27$&$\mathbf{81.19 \pm 0.34}$ \\ 
Wide-ResNet-28-10&$83.40 \pm 0.16$&$\mathbf{83.77 \pm 0.29}$&$83.53 \pm 0.16$&$84.11 \pm 0.24$&$83.09 \pm 0.30$&$\mathbf{84.89 \pm 0.12}$ \\ 
\hline SVHNcore & &  &  &  & &\\ 
Wide-ResNet-28-10&$98.05 \pm 0.03$&$98.04 \pm 0.08$ &$\mathbf{98.15 \pm 0.09}$&$\mathbf{98.07 \pm 0.03}$& $97.93 \pm 0.03$ &$98.01 \pm 0.04$ \\ 
\hline  
 \end{tabular}}
 \caption{Image Classification accuracy for WideResnet architectures~\citep{wide-resnet} trained using different augmentation policies, defined on standard~\citep{Auto-augment} and wider~\citep{trivial-augment} augmentation search spaces. We include state-of-the-art methods TA~\citep{trivial-augment}, DeepAA~\citep{deep-aa}, and 95\% confidence intervals computed over five independent runs.}\label{table:results}
 \end{table*}

\section{Experiments}\label{experiments}

\subsection{Automatic Data Augmentation}

This section showcases Algorithm~\ref{algo-pd} in common image classification benchmarks. We compare it to state-of-the-art data augmentation methods in terms of classification accuracy. Furthermore, we discuss other advantageous properties of our method through ablations. Namely, we focus on the ability to control the effect of data augmentation by modifying the constraint level, the informativeness of dual variables, and the benefits of adapting the augmentation distribution throughout training. We follow the setup~(including the transformation sets) used in recent automatic data augmentation literature~\citep{trivial-augment}. A complete list of transformations together with other hyperparameters and training settings can be found on Appendix~\ref{a:aug-details}. Note that four out of the sixteen transformations used are non-differentiable. Whereas other works have introduced gradient approximations for transformation operations with discrete parameters~\citep{bi-level-dada-augmentation, faster-auto-augment}, our approach does not require such approximations.

Throughout these experiments, we fixed the number of steps of the MH sampler~(Algorithm~\ref{algo-mh}) to two, which has the added advantage of reducing the computational cost of evaluating proposals. That is, the added computational cost of running our algorithm with respect to uniform augmentation at each iteration is essentially that of computing an additional forward pass through the neural network (see Appendix~\ref{a:runtime} for a runtime analysis). The constraint level was determined by a grid search targeting cross-validation accuracy. As shown in Table~\ref{table:results}, in both transformation sets considered, we find that our approach improves or closely matches existing approaches.

The failure to achieve large improvements in accuracy over baselines, which has been attributed to a stagnation in data augmentation research~\citep{trivial-augment}, can also reflect the limits of the benchmarking setup. In order to investigate whether larger gains can be obtained when less data is available and tasks are more challenging, we test our approach on subsampled versions of the Imagenet-100 dataset. That is, we consider all of the classes in Imagenet-100 but use only a fraction of all the training samples in the dataset to train the model. As sown in Table~\ref{table:results-imnet}, our method shows larger performance gains in this scenario, regardless of the size of the training set.

\begin{table}[]
\centering
\setlength{\tabcolsep}{6pt}
\renewcommand{\arraystretch}{1.0}
\scalebox{0.95}{
\begin{tabular}{l|cc}
\toprule
\multicolumn{1}{l|}{Samples} & TA & \textbf{OURS} \\
\hline 0.01& $19.17 \pm 0.38$& $\mathbf{20.11 \pm 0.69}$ \\ 
0.025& $34.67 \pm 0.34$& $\mathbf{37.85 \pm 0.23}$ \\ 
0.05& $48.77 \pm 0.46$& $\mathbf{51.01 \pm 1.74}$ \\ 
0.1& $60.77 \pm 0.47$& $\mathbf{62.45 \pm 0.58}$ \\ 
\hline  
 \end{tabular}
}
\caption{Image Classification accuracy for subsampled ImageNet-100 using Resnet50~\citep{resnet} and wide~\citep{trivial-augment} augmentation search space. We vary the fraction of samples kept from the original training set (first column). We include TA~\citep{trivial-augment} as a baseline and standard deviation computed over three independent runs. }\label{table:results-imnet}
\end{table}
 
\begin{figure*}[hpt!]
    \centering
     \includegraphics[width=.8\textwidth]{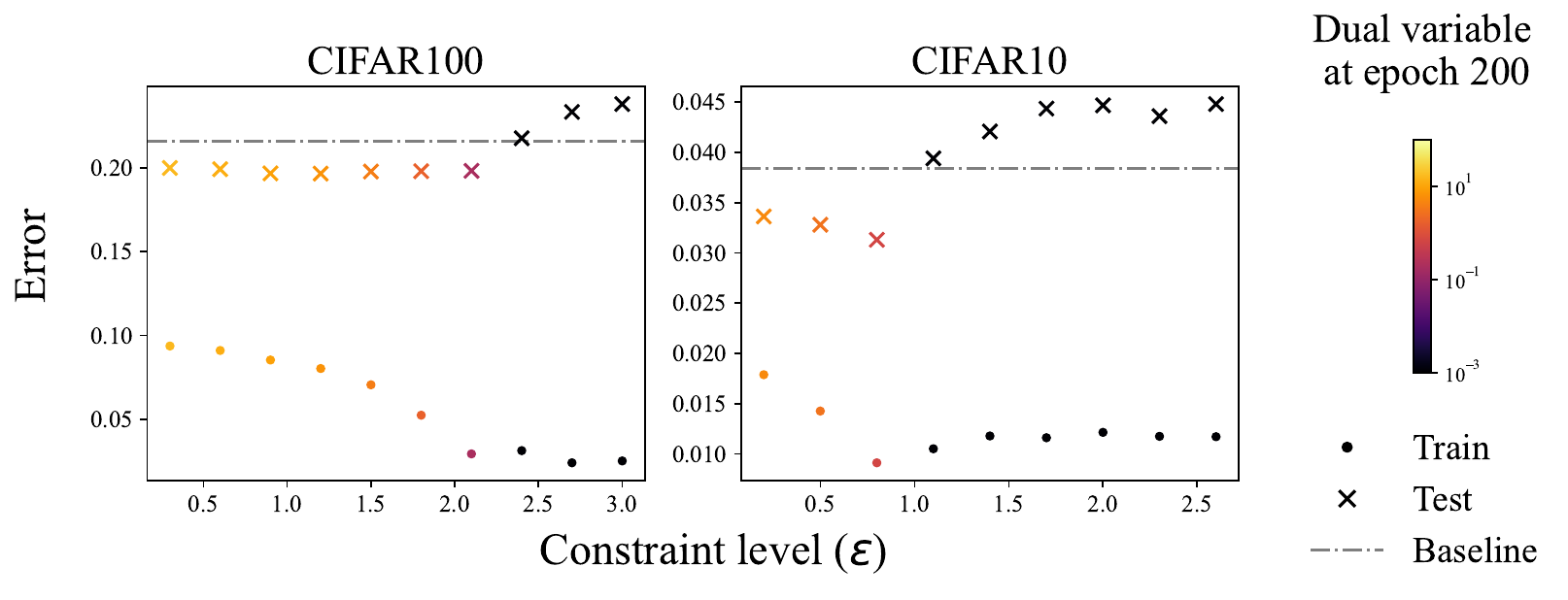}
        \caption{Constraint level ablation for WideResnet-40-2 in CIFAR datasets. We plot error rates computed over the train and test set and averaged over five runs, for different constraint levels. We include the test error of the unconstrained baseline (without augmentation) for comparison. The color of markers denotes the final value of the dual variable.}\label{plot:const-level}
\end{figure*}

 Moreover, our approach yields improvements in test accuracy over a baseline model without augmentation for a wide range of constraint levels~(Figure~\ref{plot:const-level}). This illustrates the robustness of the solution to this hyperparameter. Observe also that as the constraint is relaxed~(by increasing~$\epsilon$), the training error decreases while the generalization gap, i.e., the difference between train and test errors, increases. In other words, by loosening the invariance requirement the model can fit better to training samples at the cost of worse generalization. Eventually, only the generalization gap increases while the training error stagnates~($\epsilon > 2.1$ for CIFAR100 and $\epsilon > 0.8$ for CIFAR10). This transition occurs at the same point at which the final value of the dual variable~$\gamma$ from~(\ref{D-CERM}) essentially vanishes~(darker color). This showcases the infromativeness of the dual variable. In the case of CIFAR10, even the training error begins to increase at that point, suggesting that the invariance requirement need not be at odds with accuracy.

\begin{figure*}[hpb!]
    \centering
     \includegraphics[width=.8\textwidth]{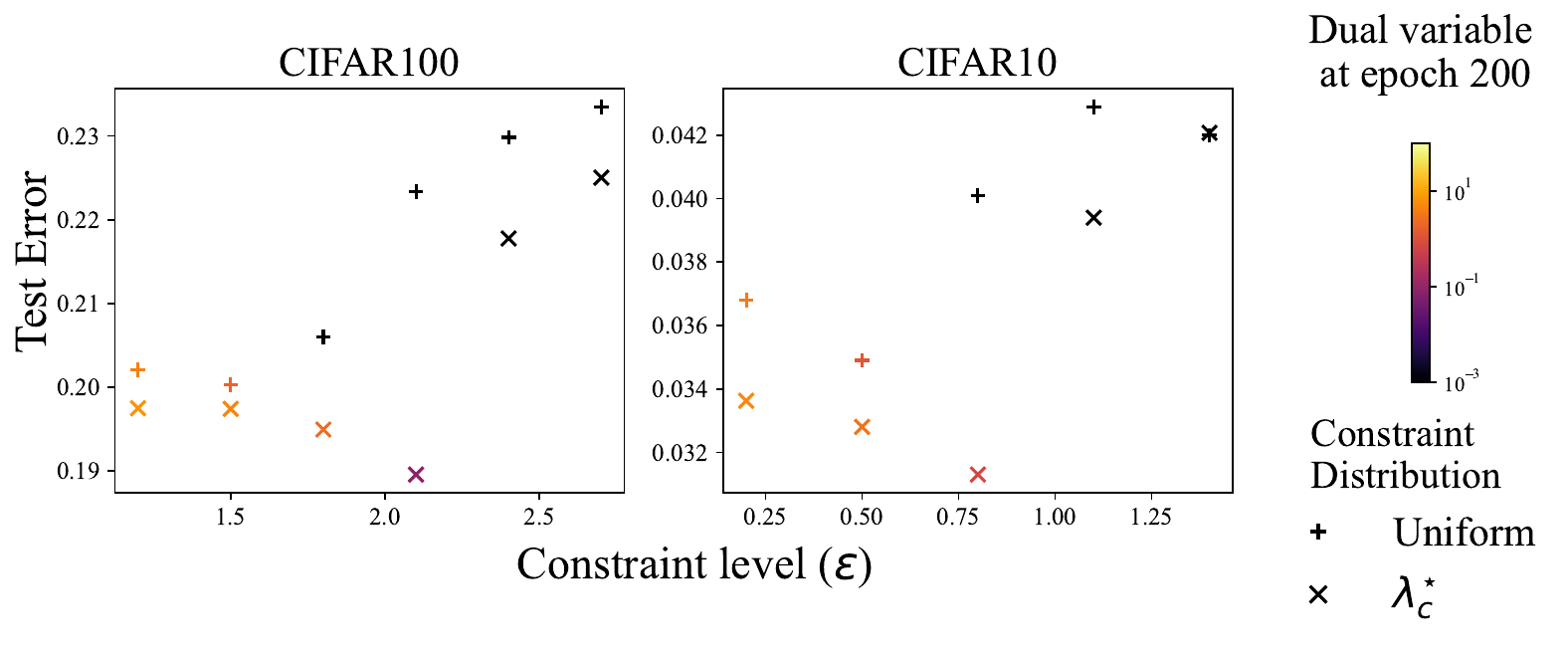}
        \caption{We compare our approach to a constraint on the uniform distribution, for WideResnet-40-2 in CIFAR datasets, at different constraint levels. We plot error rates computed over test set and averaged over five runs. Markers denote the augmentation distribution. The color of markers denotes the final value of the dual variable.}\label{plot:uniform}
\end{figure*}

Not only does Algorithm~\ref{algo-pd} tune the effect of data augmentation on the solution~(by adapting~$\gamma$ in step~9), but also modifies the distribution over transformations during training~(step~4). To showcase the benefits of this over the use of a fixed distribution, Figure~\ref{plot:uniform} compares the results obtained using our approach~(sampling according to~$\lambda^{\star}_c$) and one where step~4 is replaced by a uniform sampling over transformations. For the same constraint levels, lower test errors are obtained by sampling transformations according to $\lambda^{\star}_c$, i.e., promoting invariance. Note also that for~$\epsilon=2.1$ in CIFAR100 and $\epsilon=0.8$ for CIFAR10, the performance gap is quite large. Once again, this occurs at the point in which~$\gamma$ vanishes~(darker color) for the uniform distribution, i.e., no data augmentation occurs by the end of training. At this point, however, there is still value in promoting invariance by sampling from~$\lambda^{\star}_c$ as evidenced by the positive value of the dual variable~(lighter color) in this approach.

\begin{figure}[]
    \centering
     \includegraphics[width=.5\textwidth]{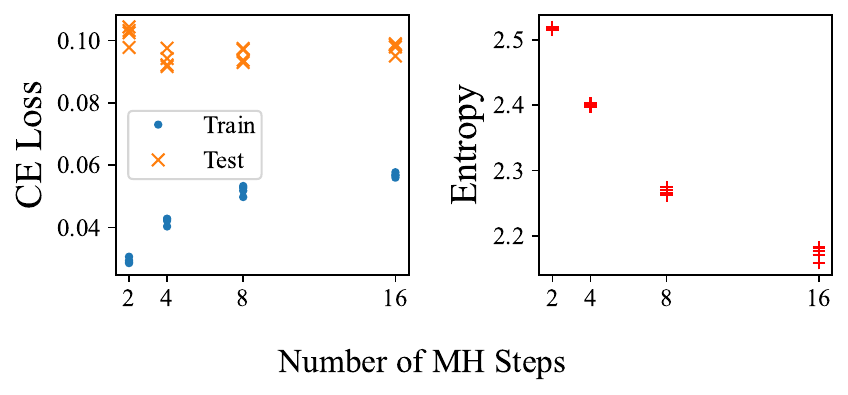}
        \caption{Number of Metropolis Hastings steps ablation for WideResnet-40-2 in CIFAR-10. The constraint level is fixed ($\epsilon=0.8$). The first plot shows the cross-entropy loss averaged over the train and test sets. The second plot shows the entropy of the augmentations sampled at the last epoch of training.  Each point represents an independent run.}\label{plot:daug:steps-ablation}
\end{figure}

Furthermore, to assess the impact of the sampling approximation on the performance of the solution we conduct an ablation on the number of steps of the MH sampler~(Algorithm~\ref{algo-mh}), keeping the constraint level ($\epsilon$) fixed. Using more steps of the chain allows samples to deviate further from the uniform distribution, which as shown in Figure~\ref{plot:daug:steps-ablation} is reflected on the lower entropy of sampled transformations (right). As shown in the left plot, this results in higher loss on the training set (represented by blue circular markers). However,  we find that increasing the number of steps does not affect significantly test loss (denoted by orange crosses). In Appendix~\ref{a:sampling-steps} we show how the number of sampling steps affects the evolution of dual variables. In Appendix~\ref{sampled-transformations} we also analyze how it affects the frequency and strength of different transformations. 

\begin{table*}[hpb!]
\centering
\setlength{\tabcolsep}{6pt} 
\renewcommand{\arraystretch}{0.9}
\scalebox{0.95}{
\begin{tabular}{@{}lll|rrrr@{}}
\hline
 &&&\multicolumn{4}{|c}{Synthetic Invariance} \\
  Dataset&Architecture&{$\gamma$}&\multicolumn{1}{r}{Full Rot.}&\multicolumn{1}{r}{Partial Rot.}&\multicolumn{1}{r}{Translation}&\multicolumn{1}{r}{Scale} \\ \hline
MNIST   & MLP & Rotation    & \textbf{0.000} & \textbf{0.004} & 3.224          & 0.035          \\
        &     & Translation & 1.344          & 0.038          & \textbf{0.289} & 0.032          \\
        &     & Scale       & 1.800          & 0.045          & 4.206          & \textbf{0.004} \\ \cline{2-7} 
        & CNN & Rotation    & \textbf{0.000} & \textbf{0.002} & 2.724          & 0.012          \\
        &     & Translation & 1.218          & 0.009          & \textbf{0.439} & 0.006          \\
        &     & Scale       & 2.026          & 0.049          & 4.029          & \textbf{0.003} \\ \hline
F-MNIST & MLP & Rotation    & \textbf{0.000} & \textbf{0.037} & 4.470          & 1.599          \\
        &     & Translation & 3.572          & 1.934          & \textbf{0.939} & \textbf{0.717}          \\
        &     & Scale       & 4.144          & 2.653          & 3.472          & 0.754 \\ \cline{2-7} 
        & CNN & Rotation    & \textbf{0.000} & \textbf{0.107} & 3.301          & 1.352          \\
        &     & Translation & 3.572          & 1.426          & \textbf{0.515} & \textbf{0.441}          \\
        &     & Scale       & 4.144          & 2.332          & 2.725          & 0.904 \\ \hline
\end{tabular}}
\caption{Value of dual variables (after 400 epochs) for different transformation constraints and synthetic invariant datasets. Columns correspond to different transformations of the dataset, and rows to dual variables associated with different transformations. Except for scaling in FashionMNIST, for all architectures and datasets the dual variable associated with the constraint corresponding to the transformations applied to the dataset is considerably lower. The smallest dual variable for each dataset and architecture is bolded.}\label{table:synthetic-invariance-results}
\end{table*}

\subsection{Synthetic Invariances}
Although our approach does not aim to learn the set of symmetries or invariant transformations from the data, but rather to impose it on the predictor, dual variables can be used to gather insights on the actual invariances underlying a learning task. We showcase this on datasets with artificial invariances, following the setup of~\cite{invariance_laplace}. Namely, we apply rotations, translations or scalings, independently drawn from the uniform distributions, to each sample in the MNIST~\citep{mnist} and FashionMNIST~\citep{fmnist} datasets. We use the same MLP and CNN architectures and hyperparameters as~\citep{invariance_laplace}. The distributions used to generate the datasets and additional experimental details can be found in Appendix~\ref{a:exp-det-synth}.

We run our algorithm constraining the loss on transformation spaces which (except for the fully rotated dataset) are larger than the true transformation range used to construct the synthetic dataset. Note that we use the same transformation sets and constraint levels ($\epsilon$) for all synthetic datasets, regardless of which transformations were used to construct the dataset. As shown in table~\ref{table:synthetic-invariance-results}, except for scalings in FashionMNIST, the dual variables ($\gamma$) associated with transformations corresponding to the true synthetic invariances in the dataset are considerably smaller. This indicates that when the transformations in the constraint correspond to a true invariance of the dataset, the constraint is easier to satisfy. 
 Therefore dual variables can give information about the invariances in a dataset or alert about the misspecification of transformation sets and constraint levels. Appendix~\ref{a:epsilon} illustrates how dual variables can be leveraged to adjust the constraint specification ($\epsilon$) using simple heuristics.

\section{Conclusion}
In this paper, we have proposed a constrained learning approach for automatic data augmentation, which instead of using augmented samples as a modified learning objective, imposes an invariance-constraint. We have shown that this yields an augmentation distribution that adapts during training, and found that coarse sampling approximations based on MCMC methods exhibit competitive performance in small and medium scale benchmarks. Furthermore, our experiments showed that dual variables can give insights about the resulting augmentation distribution. We also found that strictly feasible solutions were obtained for a wide range of constraint levels, with different generalization gaps, and that in some cases tightening the constraint even led to a lower training error. Analysing the interplay between the learning problem and the optimization algorithm, 
 and evaluating our approach in more complex or larger scale datasets, as well as in batch-mode augmentation, are promising future work directions. Finally, constrained optimization can enable further theoretical developments in data augmentation.
\newpage
\section*{Acknowledgements.}
The work of A. Ribeiro and I. Hounie is supported by NSF-Simons MoDL, Award 2031985, NSF AI Institutes program, Award 2112665, and NSF HDR TRipods Award 1934960. The work of Dr. Chamon is supported by the Deutsche Forschungsgemeinschaft (DFG, German Research Foundation) under Germany's Excellence Strategy (EXC 2075-390740016).
\bibliographystyle{apalike}
\section*{}
\bibliography{references}

\begin{thebibliography}{}

\bibitem[Benton et~al., 2020]{augerino-finzi}
Benton, G., Finzi, M., Izmailov, P., and Wilson, A.~G. (2020).
\newblock Learning invariances in neural networks from training data.
\newblock {\em Advances in Neural Information Processing Systems},
  33:17605--17616.

\bibitem[Bertsekas, 2015]{berstekas}
Bertsekas, D. (2015).
\newblock {\em Convex optimization algorithms}.
\newblock Athena Scientific.

\bibitem[Bietti and Mairal, 2019]{bietti-group-invariance}
Bietti, A. and Mairal, J. (2019).
\newblock Group invariance, stability to deformations, and complexity of deep
  convolutional representations.
\newblock {\em The Journal of Machine Learning Research}, 20(1):876--924.

\bibitem[Blaas et~al., 2021]{trua-adversarial-aug}
Blaas, A., Suau, X., Ramapuram, J., Apostoloff, N., and Zappella, L. (2021).
\newblock Challenges of adversarial image augmentations.
\newblock In {\em I (Still) Can't Believe It's Not Better! NeurIPS 2021
  Workshop}.

\bibitem[Bubeck et~al., 2015]{Langevin}
Bubeck, S., Eldan, R., and Lehec, J. (2015).
\newblock Finite-time analysis of projected langevin monte carlo.
\newblock In Cortes, C., Lawrence, N., Lee, D., Sugiyama, M., and Garnett, R.,
  editors, {\em Advances in Neural Information Processing Systems}, volume~28.
  Curran Associates, Inc.

\bibitem[Caruana et~al., 2000]{early-stopping-caruana-nn}
Caruana, R., Lawrence, S., and Giles, C. (2000).
\newblock Overfitting in neural nets: Backpropagation, conjugate gradient, and
  early stopping.
\newblock In Leen, T., Dietterich, T., and Tresp, V., editors, {\em Advances in
  Neural Information Processing Systems}, volume~13. MIT Press.

\bibitem[Cataltepe et~al., 1999]{no-free-lunch-early-stopping}
Cataltepe, Z., Abu-Mostafa, Y.~S., and Magdon-Ismail, M. (1999).
\newblock No free lunch for early stopping.
\newblock {\em Neural Computation}, 11(4):995--1009.

\bibitem[Cervino et~al., 2022]{stable-gnns}
Cervino, J., Ruiz, L., and Ribeiro, A. (2022).
\newblock Training stable graph neural networks through constrained learning.
\newblock In {\em ICASSP 2022 - 2022 IEEE International Conference on
  Acoustics, Speech and Signal Processing (ICASSP)}, pages 4223--4227.

\bibitem[Chamon et~al., 2021]{non-convex}
Chamon, L. F.~O., Paternain, S., Calvo{-}Fullana, M., and Ribeiro, A. (2021).
\newblock Constrained learning with non-convex losses.
\newblock {\em CoRR}, abs/2103.05134.

\bibitem[Chamon and Ribeiro, 2020]{Chamon-PAC}
Chamon, L. F.~O. and Ribeiro, A. (2020).
\newblock Probably approximately correct constrained learning.
\newblock {\em CoRR}, abs/2006.05487.

\bibitem[Chapelle et~al., 2000]{vicinal-risk-minimization}
Chapelle, O., Weston, J., Bottou, L., and Vapnik, V. (2000).
\newblock Vicinal risk minimization.
\newblock {\em Advances in neural information processing systems}, 13.

\bibitem[Chatzipantazis et~al., 2021]{trm-vagelis}
Chatzipantazis, E., Pertigkiozoglou, S., Dobriban, E., and Daniilidis, K.
  (2021).
\newblock Learning augmentation distributions using transformed risk
  minimization.
\newblock {\em arXiv preprint arXiv:2111.08190}.

\bibitem[Chen et~al., 2019]{Dobri-data-aug}
Chen, S., Dobriban, E., and Lee, J.~H. (2019).
\newblock A group-theoretic framework for data augmentation.

\bibitem[Cheung and Yeung, 2022]{ada-aug}
Cheung, T.-H. and Yeung, D.-Y. (2022).
\newblock Adaaug: Learning class- and instance-adaptive data augmentation
  policies.
\newblock In {\em International Conference on Learning Representations}.

\bibitem[Cohen and Welling, 2016]{Taco-OG}
Cohen, T. and Welling, M. (2016).
\newblock Group equivariant convolutional networks.
\newblock In Balcan, M.~F. and Weinberger, K.~Q., editors, {\em Proceedings of
  The 33rd International Conference on Machine Learning}, volume~48 of {\em
  Proceedings of Machine Learning Research}, pages 2990--2999, New York, New
  York, USA. PMLR.

\bibitem[Cubuk et~al., 2018]{Auto-augment}
Cubuk, E.~D., Zoph, B., Man{\'{e}}, D., Vasudevan, V., and Le, Q.~V. (2018).
\newblock Autoaugment: Learning augmentation policies from data.
\newblock {\em CoRR}, abs/1805.09501.

\bibitem[Cubuk et~al., 2020]{rand-augment}
Cubuk, E.~D., Zoph, B., Shlens, J., and Le, Q.~V. (2020).
\newblock Randaugment: Practical automated data augmentation with a reduced
  search space.
\newblock In {\em Proceedings of the IEEE/CVF Conference on Computer Vision and
  Pattern Recognition (CVPR) Workshops}.

\bibitem[Dellaportas and Roberts, 2003]{MCMC-intro}
Dellaportas, P. and Roberts, G.~O. (2003).
\newblock An introduction to mcmc.
\newblock In {\em Spatial statistics and computational methods}, pages 1--41.
  Springer.

\bibitem[DeVries and Taylor, 2017]{cutout}
DeVries, T. and Taylor, G.~W. (2017).
\newblock Improved regularization of convolutional neural networks with cutout.
\newblock {\em arXiv preprint arXiv:1708.04552}.

\bibitem[Duvenaud et~al., 2016]{early-stopping-bayesian}
Duvenaud, D., Maclaurin, D., and Adams, R. (2016).
\newblock Early stopping as nonparametric variational inference.
\newblock In Gretton, A. and Robert, C.~C., editors, {\em Proceedings of the
  19th International Conference on Artificial Intelligence and Statistics},
  volume~51 of {\em Proceedings of Machine Learning Research}, pages
  1070--1077, Cadiz, Spain. PMLR.

\bibitem[Elenter et~al., 2022]{Elenter}
Elenter, J., Naderializadeh, N., and Ribeiro, A. (2022).
\newblock A lagrangian duality approach to active learning.
\newblock {\em ArXiv}, abs/2202.04108.

\bibitem[Engstrom et~al., 2017]{madry-spatial-robustness}
Engstrom, L., Tran, B., Tsipras, D., Schmidt, L., and Madry, A. (2017).
\newblock Exploring the landscape of spatial robustness.

\bibitem[Esteves et~al., 2017]{spherical-danilidis}
Esteves, C., Allen{-}Blanchette, C., Makadia, A., and Daniilidis, K. (2017).
\newblock 3d object classification and retrieval with spherical cnns.
\newblock {\em CoRR}, abs/1711.06721.

\bibitem[Finzi et~al., 2020]{finzi-lie}
Finzi, M., Stanton, S., Izmailov, P., and Wilson, A.~G. (2020).
\newblock Generalizing convolutional neural networks for equivariance to lie
  groups on arbitrary continuous data.

\bibitem[Fioretto et~al., 2020]{Fioretto2020LagrangianDF}
Fioretto, F., Hentenryck, P.~V., Mak, T.~W., Tran, C.~D., Baldo, F., and
  Lombardi, M. (2020).
\newblock Lagrangian duality for constrained deep learning.
\newblock In {\em ECML/PKDD}.

\bibitem[Fukushima and Miyake, 1982]{fukushima}
Fukushima, K. and Miyake, S. (1982).
\newblock Neocognitron: A self-organizing neural network model for a mechanism
  of visual pattern recognition.
\newblock In {\em Competition and cooperation in neural nets}, pages 267--285.
  Springer.

\bibitem[Hastings, 1970]{MCMC-og}
Hastings, W.~K. (1970).
\newblock {Monte Carlo sampling methods using Markov chains and their
  applications}.
\newblock {\em Biometrika}, 57(1):97--109.

\bibitem[Hataya et~al., 2020]{faster-auto-augment}
Hataya, R., Zdenek, J., Yoshizoe, K., and Nakayama, H. (2020).
\newblock Faster autoaugment: Learning augmentation strategies using
  backpropagation.
\newblock In {\em European Conference on Computer Vision}, pages 1--16.
  Springer.

\bibitem[He et~al., 2015]{resnet}
He, K., Zhang, X., Ren, S., and Sun, J. (2015).
\newblock Deep residual learning for image recognition.

\bibitem[Ho et~al., 2019]{population-based-augmentation}
Ho, D., Liang, E., Chen, X., Stoica, I., and Abbeel, P. (2019).
\newblock Population based augmentation: Efficient learning of augmentation
  policy schedules.
\newblock In {\em International Conference on Machine Learning}, pages
  2731--2741. PMLR.

\bibitem[Hoffer et~al., 2020]{batch-augmentation}
Hoffer, E., Ben-Nun, T., Hubara, I., Giladi, N., Hoefler, T., and Soudry, D.
  (2020).
\newblock Augment your batch: Improving generalization through instance
  repetition.
\newblock In {\em Proceedings of the IEEE/CVF Conference on Computer Vision and
  Pattern Recognition}, pages 8129--8138.

\bibitem[Holden et~al., 2009]{adaptive-ind-mh}
Holden, L., Hauge, R., and Holden, M. (2009).
\newblock {Adaptive independent Metropolis–Hastings}.
\newblock {\em The Annals of Applied Probability}, 19(1):395 -- 413.

\bibitem[Homem-de Mello and Bayraksan, 2014]{sampling-review-stochastic-optim}
Homem-de Mello, T. and Bayraksan, G. (2014).
\newblock Monte carlo sampling-based methods for stochastic optimization.
\newblock {\em Surveys in Operations Research and Management Science},
  19(1):56--85.

\bibitem[Hornik et~al., 1989]{MLP-Universal-approx}
Hornik, K., Stinchcombe, M., and White, H. (1989).
\newblock Multilayer feedforward networks are universal approximators.
\newblock {\em Neural networks}, 2(5):359--366.

\bibitem[Immer et~al., 2022]{invariance_laplace}
Immer, A., van~der Ouderaa, T. F.~A., Fortuin, V., R{\"{a}}tsch, G., and
  van~der Wilk, M. (2022).
\newblock Invariance learning in deep neural networks with differentiable
  laplace approximations.
\newblock {\em CoRR}, abs/2202.10638.

\bibitem[Jebara, 2003]{Convex-Invariance-Learning}
Jebara, T. (2003).
\newblock Convex invariance learning.
\newblock In Bishop, C.~M. and Frey, B.~J., editors, {\em Proceedings of the
  Ninth International Workshop on Artificial Intelligence and Statistics},
  volume~R4 of {\em Proceedings of Machine Learning Research}, pages 149--156.
  PMLR.
\newblock Reissued by PMLR on 01 April 2021.

\bibitem[Ji et~al., 2021]{early-stopping-consistent}
Ji, Z., Li, J.~D., and Telgarsky, M. (2021).
\newblock Early-stopped neural networks are consistent.
\newblock In Beygelzimer, A., Dauphin, Y., Liang, P., and Vaughan, J.~W.,
  editors, {\em Advances in Neural Information Processing Systems}.

\bibitem[Joshi et~al., 2019]{semantic-adversarial}
Joshi, A., Mukherjee, A., Sarkar, S., and Hegde, C. (2019).
\newblock Semantic adversarial attacks: Parametric transformations that fool
  deep classifiers.

\bibitem[Jumper et~al., 2021]{alphafold}
Jumper, J.~M., Evans, R., Pritzel, A., Green, T., Figurnov, M., Ronneberger,
  O., Tunyasuvunakool, K., Bates, R., Z{\'i}dek, A., Potapenko, A., Bridgland,
  A., Meyer, C., Kohl, S. A.~A., Ballard, A., Cowie, A., Romera-Paredes, B.,
  Nikolov, S., Jain, R., Adler, J., Back, T., Petersen, S., Reiman, D.~A.,
  Clancy, E., Zielinski, M., Steinegger, M., Pacholska, M., Berghammer, T.,
  Bodenstein, S., Silver, D., Vinyals, O., Senior, A.~W., Kavukcuoglu, K.,
  Kohli, P., and Hassabis, D. (2021).
\newblock Highly accurate protein structure prediction with alphafold.
\newblock {\em Nature}, 596:583 -- 589.

\bibitem[K.~J.~Arrow and Uzawa., 1958]{Uzawa-alternate}
K.~J.~Arrow, L.~H. and Uzawa., H. (1958).
\newblock {\em Studies in linear and non-linear programming}.
\newblock Stanford University Press.

\bibitem[Kanbak et~al., 2017]{Geometric-robustness}
Kanbak, C., Moosavi-Dezfooli, S.-M., and Frossard, P. (2017).
\newblock Geometric robustness of deep networks: analysis and improvement.

\bibitem[Kingma and Ba, 2014]{adam}
Kingma, D.~P. and Ba, J. (2014).
\newblock Adam: A method for stochastic optimization.
\newblock {\em arXiv preprint arXiv:1412.6980}.

\bibitem[Kondor and Trivedi, 2018]{necessary-nn}
Kondor, R. and Trivedi, S. (2018).
\newblock On the generalization of equivariance and convolution in neural
  networks to the action of compact groups.

\bibitem[Krizhevsky, 2009]{CIFAR10}
Krizhevsky, A. (2009).
\newblock Learning multiple layers of features from tiny images.
\newblock Technical report, Canadian Institute for Advanced Research.

\bibitem[LeCun et~al., 2010]{mnist}
LeCun, Y., Cortes, C., and Burges, C. (2010).
\newblock Mnist handwritten digit database.
\newblock {\em ATT Labs [Online]. Available: http://yann.lecun.com/exdb/mnist},
  2.

\bibitem[Li et~al., 2020a]{early-stopping-provably-robust}
Li, M., Soltanolkotabi, M., and Oymak, S. (2020a).
\newblock Gradient descent with early stopping is provably robust to label
  noise for overparameterized neural networks.
\newblock In {\em International conference on artificial intelligence and
  statistics}, pages 4313--4324. PMLR.

\bibitem[Li et~al., 2020b]{bi-level-dada-augmentation}
Li, Y., Hu, G., Wang, Y., Hospedales, T., Robertson, N.~M., and Yang, Y.
  (2020b).
\newblock Dada: differentiable automatic data augmentation.
\newblock {\em arXiv preprint arXiv:2003.03780}.

\bibitem[Lim et~al., 2019]{fast-auto-augment}
Lim, S., Kim, I., Kim, T., Kim, C., and Kim, S. (2019).
\newblock Fast autoaugment.
\newblock {\em Advances in Neural Information Processing Systems}, 32.

\bibitem[LingChen et~al., 2020]{uniform-augment}
LingChen, T.~C., Khonsari, A., Lashkari, A., Nazari, M.~R., Sambee, J.~S., and
  Nascimento, M.~A. (2020).
\newblock Uniformaugment: A search-free probabilistic data augmentation
  approach.
\newblock {\em ArXiv}, abs/2003.14348.

\bibitem[Liu et~al., 2021]{Differentiable-aug-direct}
Liu, A., Huang, Z., Huang, Z., and Wang, N. (2021).
\newblock Direct differentiable augmentation search.
\newblock {\em CoRR}, abs/2104.04282.

\bibitem[Lyle et~al., 2020]{Gal-benefits-of-invariance}
Lyle, C., van~der Wilk, M., Kwiatkowska, M., Gal, Y., and Bloem-Reddy, B.
  (2020).
\newblock On the benefits of invariance in neural networks.

\bibitem[Madry et~al., 2017]{madry-adversarial}
Madry, A., Makelov, A., Schmidt, L., Tsipras, D., and Vladu, A. (2017).
\newblock Towards deep learning models resistant to adversarial attacks.
\newblock {\em arXiv preprint arXiv:1706.06083}.

\bibitem[Mallat, 2016]{Mallat_Understanding_CNNs}
Mallat, S. (2016).
\newblock Understanding deep convolutional networks.
\newblock {\em Philosophical Transactions of the Royal Society A: Mathematical,
  Physical and Engineering Sciences}, 374(2065):20150203.

\bibitem[M{\"{u}}ller and Hutter, 2021]{trivial-augment}
M{\"{u}}ller, S.~G. and Hutter, F. (2021).
\newblock Trivialaugment: Tuning-free yet state-of-the-art data augmentation.
\newblock {\em CoRR}, abs/2103.10158.

\bibitem[Neal, 2011]{HMC}
Neal, R.~M. (2011).
\newblock Mcmc using hamiltonian dynamics.
\newblock {\em arXiv: Computation}, pages 139--188.

\bibitem[Netzer et~al., 2011]{svhn}
Netzer, Y., Wang, T., Coates, A., Bissacco, A., Wu, B., and Ng, A.~Y. (2011).
\newblock Reading digits in natural images with unsupervised feature learning.
\newblock In {\em Advances in Neural Information Processing Systems ({NIPS})}.

\bibitem[Olah et~al., 2020]{olah2020naturally}
Olah, C., Cammarata, N., Voss, C., Schubert, L., and Goh, G. (2020).
\newblock Naturally occurring equivariance in neural networks.
\newblock {\em Distill}.
\newblock https://distill.pub/2020/circuits/equivariance.

\bibitem[Petrini et~al., 2021]{diffeo-empirical}
Petrini, L., Favero, A., Geiger, M., and Wyart, M. (2021).
\newblock Relative stability toward diffeomorphisms indicates performance in
  deep nets.
\newblock In Beygelzimer, A., Dauphin, Y., Liang, P., and Vaughan, J.~W.,
  editors, {\em Advances in Neural Information Processing Systems}.

\bibitem[P{\"{u}}schel and Moura, 2006]{Algebraic-signal-processing}
P{\"{u}}schel, M. and Moura, J. M.~F. (2006).
\newblock Algebraic signal processing theory.
\newblock {\em CoRR}, abs/cs/0612077.

\bibitem[Rice et~al., 2020]{early-stopping-adversarial}
Rice, L., Wong, E., and Kolter, Z. (2020).
\newblock Overfitting in adversarially robust deep learning.
\newblock In III, H.~D. and Singh, A., editors, {\em Proceedings of the 37th
  International Conference on Machine Learning}, volume 119 of {\em Proceedings
  of Machine Learning Research}, pages 8093--8104. PMLR.

\bibitem[Robey et~al., 2021a]{Semi-Inf}
Robey, A., Chamon, L. F.~O., Pappas, G.~J., Hassani, H., and Ribeiro, A.
  (2021a).
\newblock Adversarial robustness with semi-infinite constrained learning.
\newblock {\em ArXiv}, abs/2110.15767.

\bibitem[Robey et~al., 2021b]{robey-model-domain}
Robey, A., Pappas, G.~J., and Hassani, H. (2021b).
\newblock Model-based domain generalization.
\newblock In Ranzato, M., Beygelzimer, A., Dauphin, Y., Liang, P., and Vaughan,
  J.~W., editors, {\em Advances in Neural Information Processing Systems},
  volume~34, pages 20210--20229. Curran Associates, Inc.

\bibitem[Rockafellar and Wets, 1998]{Rockefeller}
Rockafellar, R. and Wets, R. J.-B. (1998).
\newblock {\em Variational Analysis}.
\newblock Springer Verlag, Heidelberg, Berlin, New York.

\bibitem[Rossky et~al., 1978]{langevin-OG}
Rossky, P.~J., Doll, J.~D., and Friedman, H.~L. (1978).
\newblock Brownian dynamics as smart monte carlo simulation.
\newblock {\em The Journal of Chemical Physics}, 69(10):4628--4633.

\bibitem[Russakovsky et~al., 2015]{Imagenet}
Russakovsky, O., Deng, J., Su, H., Krause, J., Satheesh, S., Ma, S., Huang, Z.,
  Karpathy, A., Khosla, A., Bernstein, M., Berg, A.~C., and Fei-Fei, L. (2015).
\newblock {ImageNet Large Scale Visual Recognition Challenge}.
\newblock {\em International Journal of Computer Vision (IJCV)},
  115(3):211--252.

\bibitem[Sannai et~al., 2019]{group-invariance-bounds}
Sannai, A., Imaizumi, M., and Kawano, M. (2019).
\newblock Improved generalization bounds of group invariant / equivariant deep
  networks via quotient feature spaces.

\bibitem[Shao et~al., 2022]{PAC-invariance-blum}
Shao, H., Montasser, O., and Blum, A. (2022).
\newblock A theory of pac learnability under transformation invariances.

\bibitem[Shen et~al., 2022]{CLIMB}
Shen, Z., Cervino, J., Hassani, H., and Ribeiro, A. (2022).
\newblock An agnostic approach to federated learning with class imbalance.
\newblock In {\em International Conference on Learning Representations}.

\bibitem[Shorten and Khoshgoftaar, 2019]{data-augmentation-survey}
Shorten, C. and Khoshgoftaar, T.~M. (2019).
\newblock A survey on image data augmentation for deep learning.
\newblock {\em Journal of big data}, 6(1):1--48.

\bibitem[Volpi et~al., 2018]{generalizing-adv-data-aug}
Volpi, R., Namkoong, H., Sener, O., Duchi, J.~C., Murino, V., and Savarese, S.
  (2018).
\newblock Generalizing to unseen domains via adversarial data augmentation.
\newblock {\em Advances in neural information processing systems}, 31.

\bibitem[Wang et~al., 2022]{representation-alignment}
Wang, H., Huang, Z., Wu, X., and Xing, E.~P. (2022).
\newblock Toward learning robust and invariant representations with alignment
  regularization and data augmentation.

\bibitem[Weiler and Cesa, 2019]{E2-G-CNN}
Weiler, M. and Cesa, G. (2019).
\newblock General e(2)-equivariant steerable cnns.
\newblock {\em CoRR}, abs/1911.08251.

\bibitem[Xiao et~al., 2017]{fmnist}
Xiao, H., Rasul, K., and Vollgraf, R. (2017).
\newblock Fashion-mnist: a novel image dataset for benchmarking machine
  learning algorithms.

\bibitem[Xu et~al., 2021]{wemix-constrained-dataaug}
Xu, Y., Noy, A., Lin, M., Qian, Q., Hao, L., and Jin, R. (2021).
\newblock Wemix: How to better utilize data augmentation.

\bibitem[Zagoruyko and Komodakis, 2016]{wide-resnet}
Zagoruyko, S. and Komodakis, N. (2016).
\newblock Wide residual networks.
\newblock {\em CoRR}, abs/1605.07146.

\bibitem[Zhang et~al., 2022]{domain-model-2}
Zhang, H., Zhang, Y.-F., Liu, W., Weller, A., Sch\"olkopf, B., and Xing, E.~P.
  (2022).
\newblock Towards principled disentanglement for domain generalization.
\newblock In {\em Proceedings of the IEEE/CVF Conference on Computer Vision and
  Pattern Recognition (CVPR)}, pages 8024--8034.

\bibitem[Zhang et~al., 2020]{adversarial-auto-augment}
Zhang, X., Wang, Q., Zhang, J., and Zhong, Z. (2020).
\newblock Adversarial autoaugment.
\newblock In {\em International Conference on Learning Representations}.

\bibitem[Zheng et~al., 2022]{deep-aa}
Zheng, Y., Zhang, Z., Yan, S., and Zhang, M. (2022).
\newblock Deep autoaugment.
\newblock In {\em International Conference on Learning Representations}.

\bibitem[Zhou et~al., 2021a]{finn-metalearning-invariance}
Zhou, A., Knowles, T., and Finn, C. (2021a).
\newblock Meta-learning symmetries by reparameterization.
\newblock In {\em International Conference on Learning Representations}.

\bibitem[Zhou et~al., 2021b]{Metaugment}
Zhou, F., Li, J., Xie, C., Chen, F., Hong, L., Sun, R., and Li, Z. (2021b).
\newblock Metaaugment: Sample-aware data augmentation policy learning.
\newblock In {\em AAAI}.

\end{thebibliography}
\appendix
\section{Additional Related work}
\subsection{Constrained Learning and Data augmentation}
\cite{wemix-constrained-dataaug} also formulate data augmentation as a constrained learning problem. They impose a constraint on the excess  risk, i.e. the difference between the statistical risk and its optimal value, on augmented data. Thus the constraint level on the augmented risk is also determined by the data distribution, augmentations considered, and model class, and the existence of a strictly feasible point is guaranteed.
\begin{equation*}
\min _{\boldsymbol{\theta} \in \Theta } R(f_{\boldsymbol{\theta}}) \quad \text { s.t. } \quad R_{aug}\left(f_{\boldsymbol{\theta}}\right)- \min _{\hat{\boldsymbol{\theta}} \in \Theta } R_{aug}(f_{\hat{\boldsymbol{\theta}}}) \leq \epsilon, \; \epsilon>0,
\end{equation*}
where $R$ and $R_{aug}$ are the statistical risk under the original and augmented distribution, exlipicity
\begin{align*}
R(f_{\boldsymbol{\theta}})=\mathbb{E}_{(\mathbf{x}, y) \sim \mathfrak{D}}[\ell(f_{\boldsymbol{\theta}}(\mathbf{x}), y)], \\  R_{\text{aug}}(f_{\boldsymbol{\theta}})= \mathbb{E}_{\substack{(\mathbf{x}, y) \sim \mathfrak{D},\\ g \sim \mathfrak{G}}}[\ell(f_{\boldsymbol{\theta}}(g\mathbf{x}), y)].
\end{align*}

Unlike our formulation, this formulation assumes a fixed distribution of augmentations $\mathfrak{G}$ is given. 

By formulating it as a constrained problem, they aim to avoid introducing a bias when the data distribution is not invariant to augmentations. Two types of biases induced by augmentation are explicitly addressed, covariate shift (i.e. label-preserving augmentations) and concept shift (i.e. label mixing augmentations). 

Interestingly, they show that under some conditions on the risk, augmented risk and constraint level, by utilizing the augmented data to constrain the solution to a small region SGD can achieve lower error \citep[Proposition 1]{wemix-constrained-dataaug}.

Instead of resorting to constrained optimization algorithms, they propose a two stage algorithm that consists of first finding an approximate minimizer of the augmented risk and then using that solution as an initialisation to the (unconstrained) statistical risk minimization problem. The first stage obtains a feasible point, and then under some conditions the SGD iterates obtained when solving the second problem remain feasible \citep[Theorem 1]{wemix-constrained-dataaug}. 

\subsection{Adversarial Data Augmentation}\label{rel-work-adv-aug}

\cite{adversarial-auto-augment} have shown that using adversarial transformations - which as already mentioned is related to promoting invariance - can give competitive results with respect to other automatic augmentation methods in image classification. However,~\cite{trua-adversarial-aug} have since evidenced the importance of two factors: the implicit learning curricula and the suboptimality of the adversarial used by~\cite{adversarial-auto-augment}, which mitigates the biases introduced by worst-case transformations. Furthermore,~\cite{trua-adversarial-aug} report that an explicit cyclic curricula in which augmentations are \emph{mild} at first, then get harder as training progresses, and finally revert to milder augmentations at the end of training, performs better empirically. We note some interesting commonalities with our approach and experimental results. First, the dynamics of our primal-dual algorithm resemble the aforementioned heuristically defined curricula. Second, the suboptimality with respect to worst case perturbations can be related to the smoothed approximation used in our approach.


\subsection{Constrained Learning and Domain Generalization}

Domain Generalization (DG) involves training the model in different but related data distributions, and evaluating in an unseen \emph{domain}. For example, a common benchmark consists of domains created by rotating images in the MNIST dataset by different angles. Constrained formulations have been proposed in this context~\citep{robey-model-domain, domain-model-2}, that enforce invariance under learnt \emph{domain translation} transformations. In contrast, our approach addresses pre-defined transformations that are commonly used in data augmentation pipelines in order to improve generalization in the same domain used to train the model.

\subsection{Group invariance}\label{a:group-stuff}
The relationship between convolutional structure and equivariance has been long known in algebraic signal processing theory \citep{Algebraic-signal-processing}. Recently, necessity results have re-gained attention in the context of neural network architecture design \citep{necessary-nn}. Several Group Convolutional neural network architectures that generalize CNNs to different groups by leveraging group convolutions have been proposed \citep{Taco-OG, spherical-danilidis,finzi-lie}. In the case of images, it has been shown that that SE2 equivariant  layers can be implemented efficiently using regular 2D convolutions~\citep{E2-G-CNN}. This approach allows to derive a parametrization for CNN filters under finite subgroups of SE2. Among other works~\citep{bietti-group-invariance, group-invariance-bounds} give theoretical analyses of the benefits of group invariance in learning settings.

As already mentioned, achieving invariance through architecture design is both challenging and limited in the sense that it relies on transformations having a specific structure (e.g: a group). The goal of exact invariance over the whole input space is more strict than the approximate invariance notion that our work addresses. As argued by \cite{Mallat_Understanding_CNNs}, CNNs can learn \emph{locally} invariant features with respect to arbitrary transformation groups, which could explain their generalization properties. Furthermore, empirical studies evidence modern CNN architectures learn approximately equivariant features to transformations such as scaling and rotations~\citep{olah2020naturally}, or diffeomorphisms~\citep{diffeo-empirical}, even when trained without direct augmentation.

However, commonly used augmentations do not form a group. Our approach does not require this structure.

\section{Additional Theoretical Details}
\subsection{Related Notions of invariance}
There are several definitions of invariance that capture different properties of the solution or data distribution under the action of transformations in $\mathcal{G}$. In the context of supervised learning, the data distribution is said to be \emph{exactly invariant} iff it does not change when transformations are applied to the covariates, i.e.,
\begin{align}~\label{inv-1}
(\mathbf{x}, y) =_d (g\mathbf{x}, y), \; \text{ for all } \; g \in \mathcal{G},
\end{align}

Note that this is equivalent to the distribution of inputs $x$ being conditionally invariant on the label $y$,
\begin{align}~\label{inv-2}
(g X \mid Y=y)={ }_d(X \mid Y=y).
\end{align}

If the data distribution $\mathfrak{D}$ is statistically invariant under the action of $\mathcal{G}$, invariant solutions have provable advantages in terms of sample complexity~\citep{Dobri-data-aug,bietti-group-invariance, group-invariance-bounds, Gal-benefits-of-invariance}.

These notions of invariance do not explicitly contemplate the task at hand, that is, not all changes in $y$ equally affect performance. Thus, we use the loss to encode meaningful differences in labels with respect to the task, as described in section~\ref{proposed:invariance-as-robustness}. Throughout this work, we thus refer to the invariant risk $R_{\text{inv}}$ defined on equation~\ref{worst-case} as the degree of invariance, unless otherwise noted.

\subsection{Data Augmentation and Penalised Formulations}\label{a:aug-expectation}

Augmented risk minimisation formulations usually do not  include the loss on clean data explicitly on the objective. 
However, several common choices for the augmentation distribution $\mathfrak{G}$ have positive mass at the identity transformation $e(\mathbf{x}) = \mathbf{x}$, i.e. $P_{\mathfrak{G}}(g=e) = \gamma>0$.

 Therefore, the risk under the augmentation distribution $\mathfrak{G}$ is statistically equivalent to a regularised objective, explicitly
\begin{align*}\label{Reg-Aug-SR}
 \mathbb{E}_{\substack{(\mathbf{x}, y) \sim \mathfrak{D},\\ g \sim \mathfrak{G}}}[\ell(h(g\mathbf{x}), y)] = & \gamma \mathbb{E}_{(\mathbf{x}, y) \sim \mathfrak{D}}[\ell(h(\mathbf{x}), y)]\\& + \mathbb{E}_{\substack{(\mathbf{x}, y) \sim \mathfrak{D},\\ g \sim \tilde{\mathfrak{G}}}}[\ell(h(g\mathbf{x}), y)],
\end{align*}
where $\tilde{\mathfrak{G}}$ is a distribution over $\mathcal{G}$ with no mass at the identity.

The bias-variance tradeoff arising from averaging  over transformations of training samples while introducing a distribution shift~\citep{Dobri-data-aug}, can thus be controlled through the probability of the identity transformation in the augmentation distribution.

\subsection{Worst case transformations as an Augmentation Distribution}\label{a:invariant-loss-lag}

In this section we show that finding the transformation that maximises the loss can be interpreted as an augmentation distribution. We thus extend the particular case of bounded additive perturbations in~\cite{Semi-Inf} to a transformation set $\mathcal{G}$.

Given a predictor $f_{\theta}$ and a data instance $(x, y)$, we want to solve the optimization problem
\begin{align}
P^* = \max_{g \in \mathcal{G}}\ell\left(f_{\boldsymbol{\theta}}\left(g\mathbf{x}\right), y\right)
\end{align}
Assuming  $\ell \in L^{2}$, we begin by writing the problem in epigraph form
\begin{align}\tag{EPI}\label{EPI}
P^* = \min_{t \in L^{2}} & \; t(g)\\
\text{s.t. }& \ell(f_{\theta}(g x), y) \leq t(g) \; \forall g \in \mathcal{G}\notag
\end{align}

Then, the Lagrangian associated to this problem can be written as
\begin{align*}
L(\lambda, t)& = t(g)+\int_{\mathcal{G}} \lambda(g)(\ell(f_{\theta}(g x), y)-t(g))dg\\
& = t(g)\left[1-\int_{\mathcal{G}} \lambda(g) dg\right] \\
& \;+ \int_{\mathcal{G}} \lambda(g)(\ell(f_{\theta}(g\mathbf{x}), y)) dg.
\end{align*}

Problem (\ref{EPI}) can thus be written in Lagrangian form as:
\begin{align*}
P^* = \sup _{\lambda \in L^{2}_{+}}\min_{t \in L^{2}} & \; L(\lambda, t),
\end{align*}
where $L^{2}_{+}$ is the space of non negative functions.

Since $L$ is linear in $t$, if $P^*$ is finite, then strong duality holds \citep[Chapter~4]{berstekas}. Then we can solve the dual problem:
\begin{equation*}
 P^* = D^* := \sup _{\lambda \in \mathcal{P}^{q}} \min _{t  \in L^{p}} L(\lambda, t),
\end{equation*}
 Since t is unconstrained, the inner minimization yields:
 $$
 \min _{t  \in L^{2}} L(\lambda, t) =
\left\{\begin{array}{l}
-\infty \quad \text{ if } \int_{\mathcal{G}} \lambda(g) dg \neq 1 \\
\int_{\mathcal{G}} \lambda(g)\ell(f_{\theta}(g\mathbf{x}), y) dg \quad \text { otherwise }
\end{array}\right.
$$
Maximizing over $\lambda$ we get the desired result:
 \begin{align*}
 P^* = & \sup_{\lambda \in L^{2}_{+}} \int_{\mathcal{G}} \lambda(\mathbf{x}, y, g)\ell(f_{\theta}(g\mathbf{x}), y)dg,\\ \nonumber
 & \text{ s. t.}  \int_{\mathcal{G}} \lambda(\mathbf{x}, y, g)dg = 1\;\;
 \end{align*}
 Since any solution of this optimization problem, which we will denote $\lambda^{\star}(g)$, is non-negative almost everywhere and integrates one, it defines a distribution over the set of transformations $\mathcal{G}$. This allows us to re-interpret the maximization over $\mathcal{G}$ as an expectation, namely
 \begin{align*}
 \max_{g \in \mathcal{G}}\ell\left(f_{\boldsymbol{\theta}}\left(g\mathbf{x}\right), y\right) = \mathbb{E}_{g \sim \lambda^{\star}}\left[ \ell(f_{\boldsymbol{\theta}}(g\mathbf{x}), y) \right].
 \end{align*}
\subsection{Derivation of the smoothed distribution $\lambda^{\star}_c$}\label{a:sm-adv-loss}
In this section we show that for a particular value of $c$, the smoothed distribution $\lambda^{\star}_c$ has a closed form solution that enables MCMC sampling methods.
We defined the $\mathcal{L}^2$ regularised problem as
\begin{align*}
&\sup_{\lambda \in \mathcal{L}^2_+} \int_{\mathcal{G}} \left[\lambda(\mathbf{x}, y, g)\ell(h(g\mathbf{x}), y) + c  \lambda^2(\mathbf{x}, y, g)\right] dg .\\
& \text{ s . t. } \int_{\mathcal{G}} \lambda(\mathbf{x}, y, g) dg  = 1
\end{align*}
 
 We will begin by finding the optima of the unconstrained problem.
 
Since $\mathcal{L}^2$ is decomposable, and the integral is Careothodory, the supremum can be found by taking pointwise supremum in $\mathbb{R}_+$ inside the integral. See ~\cite{Rockefeller} chapter 14, theorem 14.60.
\begin{align*}
 \sup _{\lambda \in \mathcal{L}^{2}_+} & \int_{G} \big[ \lambda(\mathbf{x}, y, g)\ell(f_{\theta} (g x), y)\\
  & \; - c \int_{\mathcal{G}} \lambda^{2}(\mathbf{x}, y, g) \big] dg \\
  = \int_{\mathcal{G}} \sup _{\lambda(\mathbf{x}, y, g) \in \mathcal{R}_+}& \big[ \lambda(\mathbf{x}, y, g)\ell(f_{\theta} (g x), y) \\
  & \quad  \quad \quad \quad \quad \quad -c \lambda^{2}(\mathbf{x}, y, g) \big] dg
\end{align*}
Note that if $\ell(f_{\theta} (g x), y))<0$ then the pointwise maximum is attained at $\lambda^*(x,y,g) = 0$.

Otherwise, we can find the pointwise maxima by setting the partial derivative w.r.t. $\lambda$ to zero.
\begin{align*}
    \frac{\partial }{\partial \lambda(\mathbf{x}, y, g)} \left[ \lambda(\mathbf{x}, y, g)\ell(f_{\theta} (g x), y)-c \lambda^{2}(\mathbf{x}, y, g) \right] \\
    = \ell(f_{\theta} (g x), y) - 2c \lambda(\mathbf{x}, y, g) \\
\end{align*}
Then, if $\ell(f_{\theta} (g x), y))\geq 0$ the pointwise maximum is attained at $\lambda^*(x,y,g) = \frac{\ell(f_{\theta} (g x), y)}{2c}$.
Then the optimum of the unconstrained problem can be written as
\begin{align*}
\lambda^*(x,y,g) = \left[\frac{\ell(f_{\theta} (g x), y)}{2c}\right]_{+}
\end{align*}
For the solution of the unconstrained problem $\lambda^*$ to be feasible for the constrained problem, we can choose $c$ so that it satisfies the normalization constraint:
\begin{align*}
\int_{\mathcal{G}} \lambda^*(\mathbf{x}, y, g) dg  = \frac{1}{2c}\int_{\mathcal{G}} \left[\ell(f_{\theta} (g x), y)\right]_{+} dg = 1\\
\Leftrightarrow \\
c = \frac{1}{2}\int_{\mathcal{G}} \left[\ell(f_{\theta} (g x), y)\right]_{+} dg
\end{align*}
If we further assume the loss is non-negative, which holds for many commonly used losses such as cross-entropy and mean squared error, we can drop the projection into the non-negative orthant.

\FloatBarrier

\subsection{Constrained Statistical Learning Guarantees}\label{a:chamon-guarantees}

In this section we provide an overview of the results that bound the empirical duality gap of problem (\ref{D-CERM}) under uniform convergence assumptions shown by~\cite{Chamon-PAC, non-convex} in more general settings and in~\cite{Semi-Inf} for the particular case of robustness (invariance) objectives. The bound goes to show that the sample complexity of constrained learning is not larger than that of unconstrained ERM. However, unlike the unconstrained formulation of traditional data augmentation in~(\ref{Aug-ERM}), our constrained formulation has stronger guarantees over the loss of transformed samples.

\textbf{A1.} (Bounded Distance to Convex Hull) The parametrization $f_\theta$ is rich enough so that for each $\boldsymbol{\theta}_1, \boldsymbol{\theta}_2 \in \Theta \text{ and } \beta \in[0,1]$, there exists $\theta \in \Theta$ such that $\sup _{\mathbf{x} \in \mathcal{X}}\left|\beta f_{\theta_1}(\boldsymbol{x})+(1-\beta) f_{\theta_2}(\boldsymbol{x})-f_\theta(\boldsymbol{x})\right| \leq \alpha$.

\textbf{A2.}(Slater's Condition) There exists $\boldsymbol{\theta}^{\prime} \in \Theta$ such that $\mathbb{E}_{(\mathbf{x}, y) \sim \mathfrak{D}}\left[\mathbb{E}_{g \sim \lambda^{\star}}\left[ \ell(f_{\boldsymbol{\theta}}(g\mathbf{x}_n), y_n))\right]\right] - \epsilon > M\nu >0 $.

\textbf{A3.}(Uniform convergence) There exists $\zeta_0(N, \delta), \zeta_1(N, \delta) \geq 0$ monotonically decreasing in $N$ such that w.p $1-\delta$  $\forall \theta \in \Theta$:
$$
\begin{aligned}
&\left|\mathbb{E}_{(\boldsymbol{x}, y) \sim \mathfrak{D}}\left[\mathbb{E}_{g \sim \lambda^{\star}}\left[ \ell(f_{\boldsymbol{\theta}}(g\mathbf{x}), y)) \right]\right] \right.\\
&\left. \quad \quad \quad -\frac{1}{N} \sum_{n=1}^N \mathbb{E}_{g \sim \lambda^{\star}}\left[ \ell(f_{\boldsymbol{\theta}}(g\mathbf{x}_n), y_n)) \right] \right| \leq \zeta_1(N, \delta) \\
&\left|\mathbb{E}_{(\boldsymbol{x}, y) \sim \mathfrak{D}}\left[\ell\left(f_{\boldsymbol{\theta}}(\mathbf{x}), y\right)\right] \right. \\
& \left. \quad \quad -\frac{1}{N} \sum_{n=1}^N \ell\left(f_{\boldsymbol{\theta}}\left(\mathbf{x}_n\right), y_n\right)\right| \leq \zeta_0(N, \delta) .
\end{aligned}
$$
\textbf{A4.} The loss function $\ell(\cdot, y)$ is convex, $M$-Lipchitz and $[0, B]$ bounded.

The first two assumptions state that the hypothesis space $\mathcal{H}_{\boldsymbol{\theta}}=\left\{f_{\boldsymbol{\theta}} \mid \boldsymbol{\theta} \in \Theta \subseteq \mathbb{R}^{p} \right\}$ is sufficiently expressive in the sense that the pointwise distance to its convex hull $\text{conv}(\mathcal{H}_{\boldsymbol{\theta}})$ is bounded (A1), and there exists a function that $f_{\theta} \in \mathcal{F}$ that strictly satisfies the constraint (A2). Note that in the case of neural networks, universal approximation results~\cite{MLP-Universal-approx} imply the existence of hypothesis spaces that are sufficiently expressive to satisfy these assumptions. However, this may be at odds with uniform convergence (A3), which limits the complexity of the hypothesis space. 

\textbf{Empirical Duality Gap Bound}~\cite{non-convex}
Under Assumptions $A1-A4$, there exists a solution $(\gamma^*, f_{\hat{\boldsymbol{\theta}}^{\star}})$ of the empirical dual problem (\ref{D-CERM}) such that, w. p. $1-5 \delta$, it holds
\begin{align*}
\left|P^{\star}-D_{\text{emp}}^{\star}\right| \leq(1+\overline{\gamma})(M \nu+\max(\zeta_0(N, \delta), \zeta_1(N, \delta)), \text{ and }\\
\mathbb{E}_{\mathfrak{D}}\left[\mathbb{E}_{g \sim \lambda^{\star}}\left[ \ell(f_{\boldsymbol{\theta}}(g\mathbf{x}), y) \right]\right] \leq \epsilon +\zeta_1\left(N, \delta\right),
\end{align*}
where $P^{\star}$ is the value of (\ref{CSRM}), $\bar{\zeta}=\max _i \zeta_i\left(N_i, \delta\right)$, and $\overline{\gamma}$ is lower bounded by $\gamma^*$.

The optimal dual variable $\gamma^*$ depends not only on the data distribution, loss and hypothesis class, but also on the constraint level. As a result, if satisfying the invariance constraint is overly restrictive for a learning task and results in large optimal dual variables, it also leads to looser guarantees for our empirical dual algorithm.

\section{Experimental Setup Details}
\subsection{Automatic Data Augmentation}\label{a:aug-details}
\subsubsection{Transformations}
As in recent automatic augmentation literature~\citep{population-based-augmentation, fast-auto-augment, faster-auto-augment, adversarial-auto-augment, rand-augment, uniform-augment, adversarial-auto-augment, trivial-augment,Metaugment, deep-aa, ada-aug}, we focus on image classification datasets and employ a transformation search space comprising 14 operations, introduced by \cite{Auto-augment}. For those that have parameters, their magnitudes are discretized in thirty levels, which does not compromise performance and greatly reduces the search space~\citep{rand-augment}. Most approaches compose transformations, i.e. applying more than one transformation to the same image. However, recently \cite{trivial-augment} have shown that applying only one transformation at a time, defined over a wider magnitude space (noted Wide in Table \ref{Table:aug-search-space}) can outperform other approaches. We thus use the same transformation space as \citep{trivial-augment}.

Table~\ref{Table:aug-search-space} from~\cite{trivial-augment} lists the operations and their magnitude ranges. In our experiments we used the wide search space, the standard ranges from~\cite{rand-augment} are included for comparison. We extend the codebase provided by~\cite{trivial-augment}, which uses the Pillow\footnote{~\href{https://github.com/python-pillow/Pillow}{https://github.com/python-pillow/Pillow}} implementation of all transformations except for cutout, and refer to its documentation for further details about image operations.
\begin{table}[h]
    \centering
    \setlength{\tabcolsep}{6pt} 
\renewcommand{\arraystretch}{1.3}
    \begin{tabular}{l|c c}
\hline \multicolumn{1}{c|}{ \textbf{Operation} } &\multicolumn{2}{c}{ \textbf{Magnitude} }   \\
&Standard& Wide\\
\hline Identity & $-$ & $-$\\
ShearX & {$[-0.3,0.3]$} & {$[-0.99,0.99]$} \\
ShearY & {$[-0.3,0.3]$} &{$[-0.99,0.99]$} \\
TranslateX & {$[-10,10]$} & ${[-32,32]}$\\
TranslateY & {$[-0.45,0.45]$} & ${[-32,32]}$ \\
Rotate & {$[-30,30]$} &{$[-135,135]$} \\
AutoContrast & $-$ & $-$\\
Invert & $-$ & $-$\\
Equalize & $-$ & $-$\\
Solarize & {$[0,256]$} & \\
Posterize & {$[4,8]$} & {$[2,8]$}\\
Contrast & {$[0.1,1.9]$} &{$[0.01,2]$} \\
Color & {$[0.1,1.9]$} &{$[0.01,2]$} \\
Brightness & {$[0.1,1.9]$} & {$[0.01,2]$}\\
Sharpness & {$[0.1,1.9]$} & {$[0.01,2]$}\\
Flips & $-$ & $-$\\
Cutout & $16(60)$ & $16(60)$\\
Crop & $-$ & $-$\\
\hline
\end{tabular}\caption{Pillow image operations in the data augmentation search space and the range of magnitudes corresponding to the standard~\citep{rand-augment} and wide \citep{trivial-augment} search
spaces. Some operations do not use magnitude parameters.}\label{Table:aug-search-space}
\end{table}
\subsubsection{Training Setup}
In order to enable comparisons and reproducibility we use the same training pipeline as in previous works~\cite{trivial-augment} . We apply the vertical flip and
the pad-and-crop augmentations and a 16 pixel cutout~\citep{cutout} after any augmentation method. We trained Wide-ResNet~\citep{wide-resnet} models in the 40-2 and 28-10 configurations For CIFAR~\cite{CIFAR10} and SVHN~\cite{svhn} datasets, and ResNet-50~\cite{resnet} for Imagenet-100~\cite{Imagenet}. 

Except for epoch ablation and Imagenet-100 experiments, we use SGD with Nesterov Momentum and a learning rate of 0.1, a batch size of 128, a 5e-4 weight decay. In Imagenet-100 experiments, we use Adam~\cite{adam} and a learning rate of 5e-4. We use a cosine learning rate decay schedule and train for 270 epochs for tinyImagenet and 200 epochs for all other datasets.
In ablation experiments we also trained for 600 epochs and used a custom learning rate schedule. For the first 185 epochs we followed the same cosine learning rate decay schedule, and then switching to a custom step learning rate scheduler detailed on Table~\ref{Table:lr-schedule-600}. This schedule was implemented after observing that just scaling the cosine learning rate schedule to 600 epochs resulted in slow convergence, thus yielding solutions similar to training for 600 epochs, and failing reflect the effects of early stopping which this experiment addressed.

We used 2 MH steps for the MH sampler (Algorithm~\ref{algo-mh}) unless stated otherwise, and a learning rate of $10^{-3}$ for the dual ascent step.

No other hyperparameters were tuned or modified with respect to standard settings. The constraint levels were set for Wide-ResNet-40-2 in the wide augmentation space by maximising three fold cross validation over the grid specified on Table~\ref{table:constraint-level-grid-search}. Then the constraint levels were adjusted for other architectures and search spaces so that the dual variables at the end of training were small but not zero (of the order of $10^{-1}$), which empirically showed good results.  The resulting constraint levels corresponding to the results in~\ref{table:results} are detailed in table~\ref{table:constrain-level-grid}.

We used the provided code and reported hyperparameters for running TrivialAugment~\citep{trivial-augment} and DeepAA~\citep{deep-aa}. For the latter, unlike the results reported in~\citep{deep-aa}, we kept all evaluation hyperparameters (including weight decay) consistent with that of other methods. Results for the wide augmentation space were not included in~\citep{deep-aa}. We thus performed the augmentation policy search for the wide search space using the same hyperparameters (except for the augmentation space) reported in~\citep{deep-aa} for CIFAR datasets. We also run the search for both augmentation spaces in SVHN datasets, and evaluated the policy with the same setup as other methods.

\begin{table}[!h]
\centering
\setlength{\tabcolsep}{6pt} 
\renewcommand{\arraystretch}{1.3}
\begin{tabular}{c|c}
\hline
\textbf{Epochs} & \textbf{LR Scheduler Step} \\ \hline
180-230                                 & 10                \\
230-430                                 & 20                \\
430-600                                 & 40                \\ \hline
\end{tabular}\caption{Learning Rate Custom schedule used when training for 600 epochs. We use the standard scheduler for the first 180 epochs, and then update the LR only every $n$ epochs, where $n$ is the number indicated in the second column. This hand designed schedule outperforms using a cosine learning rate schedule for 600 epochs, but could improvements convergence speed and performance by exploring other LR-schedulers or tuning it.}\label{Table:lr-schedule-600}
\end{table}

\begin{table}[!h]
\centering
\setlength{\tabcolsep}{6pt} 
\renewcommand{\arraystretch}{1.3}
\begin{tabular}{l|c}
\hline
\textbf{Dataset} & \multicolumn{1}{c}{\textbf{Constraint Level Grid Range}} \\ \hline
CIFAR10  & {[}0.2, 2.3{]} \\
CIFAR100 & {[}0.3, 2.7{]} \\
SVHNcore & {[}0.1, 1.5{]} \\ \hline
\end{tabular}\caption{Constraint level grid search space. For each dataset, we evaluated three fold cross validation accuracy for 8 evenly spaced constraint levels in the ranges given in the second column. The one with the highest cross validation score was then selected and used to train the model with the full dataset.}\label{table:constraint-level-grid-search}
\end{table}

\begin{table}[!h]
\centering
\setlength{\tabcolsep}{6pt} 
\renewcommand{\arraystretch}{1.2}
\begin{tabular}{l|cc}
\hline
 & \multicolumn{2}{c}{\textbf{Constraint Level}} \\ 
    & Standard  & Wide  \\  \hline
CIFAR10  &  &\\
Wide-ResNet-40-2   &   $0.6$ & $0.8$ \\
Wide-ResNet-28-10   &   $0.4$   & $0.8$ \\\hline
CIFAR100 & & \\
Wide-ResNet-40-2    &   $0.9$ & $0.8$ \\
Wide-ResNet-28-10    &  $0.9$   &  $1.2$ \\\hline
SVHNcore & & \\
Wide-ResNet-28-10     &  $0.1$  & $0.2$ \\\hline
\end{tabular}
\caption{Constraint levels for different datasets and architectures, for the results presented in Table~\ref{table:results}}\label{table:constrain-level-grid}
\end{table}

\subsection{Syntetic Invariances}\label{a:exp-det-synth}

We showcase our approach on datasets with artificial invariances, following the setup of~\cite{invariance_laplace}. Explicitly, we generate the synthetic datasets, by applying either rotations, translations or scalings, to each sample in the MNIST~\cite{mnist} and FashionMNIST~\cite{fmnist} datasets. The transformations are sampled from uniform distributions over the ranges detailed in Table~\ref{table:synthetic-invariance-dist}. We use the same MLP and CNN architectures and hyperparameters as~\cite{invariance_laplace}, except that we use only 1 augmentation per sample instead of 31 during training.

 For each transformation set (rotations, translations and scalings), we constraint the expected loss over samples augmented with the transformations sampled from uniform distributions over the ranges detailed in Table~\ref{table:synthetic-invariance-constraint}. Note that there is a mismatch between the distribution used to generate the data and that used in the constraints. That is, except for the fully rotated dataset, the constraints are larger than the true transformation range used to construct the synthetic dataset (Table~\ref{table:synthetic-invariance-dist}). We use the same transformation sets and constraint specification ($\epsilon$) for all synthetic versions of the datasets. We only vary the constraint level depending on the architecture, to contemplate how model capacity affects the relative difficulty of satisfying the constraint. In order to do so, we evaluate the loss of a model trained without constraints. Constraint levels are specified in Table~\ref{table:synthetic-invariance-eps}. 

\begin{table}[htpb]
\setlength{\tabcolsep}{5pt} 
\renewcommand{\arraystretch}{0.9}
\centering
\scalebox{0.8}{
\begin{tabular}{l|l|l}
Synthetic invariance & Parameter              & Distribution             \\ \hline
Full Rotation & Angle in radians. & $\mathcal{U}\left[-\frac{\pi}{2}, \frac{\pi}{2}\right]$ \\
Partial Rotation     & Angle in radians.      & $\mathcal{U}[-\pi, \pi]$ \\
Translation          & Translation in pixels. & $\mathcal{U}[-8,8]^2$    \\
Scale                & Exponential Scaling factor.        & $\mathcal{U}[-log(2),log(2)]$      \\ \hline
\end{tabular}}
\caption{Sampling parameters for transformations used to obtain synthetically invariant datasets, from~\citep{invariance_laplace}.}\label{table:synthetic-invariance-dist}
\end{table}

\begin{table}[htpb]
\centering
\setlength{\tabcolsep}{6pt} 
\renewcommand{\arraystretch}{1.0}
\scalebox{0.9}{
\begin{tabular}{l|l|l}
Constraint Set & Parameter                   & Range         \\ \hline
Rotations      & Angle in radians.           & $[-\pi, \pi]$ \\
Translation    & Translation in pixels.      & $[-16,16]^2$  \\
Scale          & Exponential Scaling factor. & $[-1.5,1.5]$  \\ \hline
\end{tabular}}
\caption{Transformation sets used as invariance constraints. All sets are used simultanously, with the same constraint level ($\epsilon$) for all transformations.}\label{table:synthetic-invariance-constraint}
\end{table}

\begin{table}[]
\centering
\setlength{\tabcolsep}{6pt} 
\renewcommand{\arraystretch}{1.0}
\scalebox{0.9}{
\begin{tabular}{lll}
        & MLP & CNN  \\ \hline
MNIST   & 0.6 & 0.35 \\
F-MNIST & 0.8 & 0.6  \\ \hline
\end{tabular}
}
\caption{Constraint levels used for different architectures and base datasets. We use the same constraint levels for all synthetically invariant versions of the same base dataset.}\label{table:synthetic-invariance-eps}
\end{table}
\section{Additional Experimental Results}\label{a:results:aug}

\subsection{Automatic Data augmentation}
The following section contains additional ablations and discussions about our algorithm. As in previous sections, we use the wide~\cite{trivial-augment} augmentation space and CIFAR image classification benchmarks. Our main motivation is to analyse how the different hyperparameter choices and the learning algorithm affect the generalization and invariance of the obtained solutions. First, we show how the dual variables adapts to different constraint levels during training, and link its dynamics to heuristically defined learning curricula. We then evaluate the effect of training for more epochs and link our observations to the known properties of early stopping in unconstrained learning, in section~\ref{early-stopping}. In section~\ref{a:sampling-steps}, we analyse how the sampling approximation affects regularisation, by performing an ablation on the number of MH steps. Lastly, we include the observed frequencies of sampled transformations for different setups, so as to obtain further insights in how the distribution adapts throughout training (Section~\ref{sampled-transformations}).

\subsubsection{Dual Variable Dynamics}
\begin{figure}[h]
\centering
     \includegraphics[width=.45\textwidth]{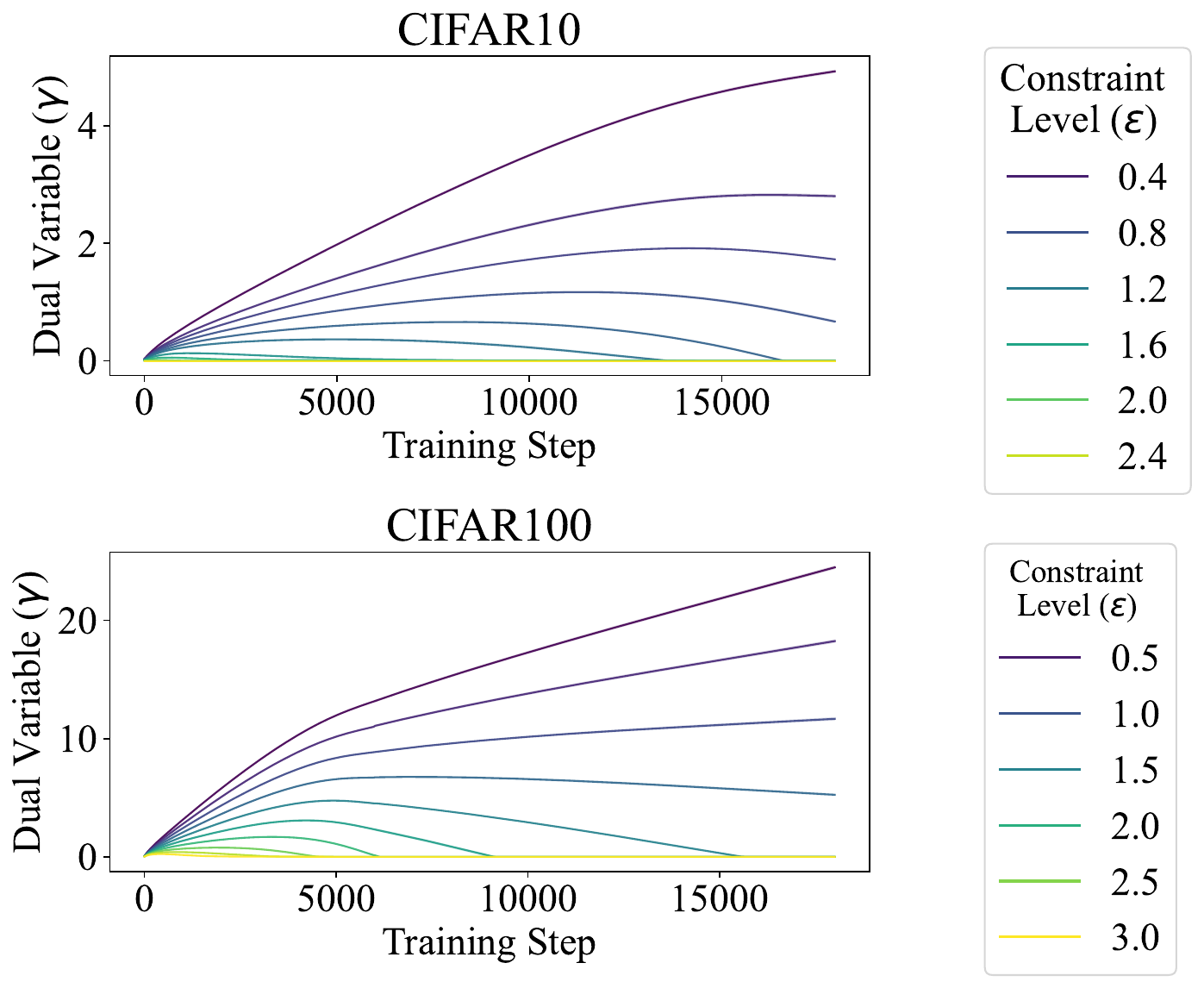}
        \caption{Evolution of dual variables during training for Wide-ResNet-40-2 in CIFAR datasets. For most levels, augmentation increases until the constraint becomes feasible and then decreases towards the end of training. For stricter levels, the algorithm does not converge in 200 epochs using the standard learning settings. However, the solutions obtained still show good properties.}\label{a:plot:dual-level}
\end{figure}

As already mentioned, the dual variables control the weight of augmented samples during training, thus balancing the trade-off between fitting the primal objective (i.e. loss over training samples) and satisfying the constraints (i.e. loss over transformed samples). In penalised approaches, on the contrary, this trade-off is fixed. In Figure~\ref{a:plot:dual-level} we show how the dual variable adapts to different constraint levels, for Wide-ResNet-40-2 in CIFAR datasets using the standard setup. Note that for stricter constraint levels, the algorithm has not converged when it reaches 200 epochs. We also observe that the dynamics of our primal-dual algorithm resembles the augmentation learning curricula proposed by~\cite{trua-adversarial-aug}.

\subsubsection{Early Stopping}\label{early-stopping}

Our approach has slower convergence than  unconstrained approaches and stopping training arbitrarily - after a fixed number of epochs - can result in solutions that are unfeasible or sub-optimal. However, early stopping is a popular regulariser, particularly for neural networks trained through gradient descent. Several empirical \citep{early-stopping-caruana-nn, early-stopping-adversarial} and theoretical~\citep{early-stopping-consistent, early-stopping-provably-robust, early-stopping-bayesian} results show its advantages in terms of generalisation and robustness to noisy training data. In general, the advantages of early stopping do not lie in the sub-optimality of the solution in terms of training error, but on nature of the regularization or prior imposed, which leads to non-uniform model selection among models with a given training error~\citep{no-free-lunch-early-stopping}.

To the best of our knowledge, there is no literature that explicitly addresses early stopping in empirical primal-dual learning. Whether the generalisation gap in constraint satisfaction can be reduced by early stopping regularisation, in the same manner early stopping regularisation can reduce the generalisation gap in unconstrained learning, thus remains unclear. 

\begin{figure}[htpb]
    \centering
     \includegraphics[width=.45\textwidth]{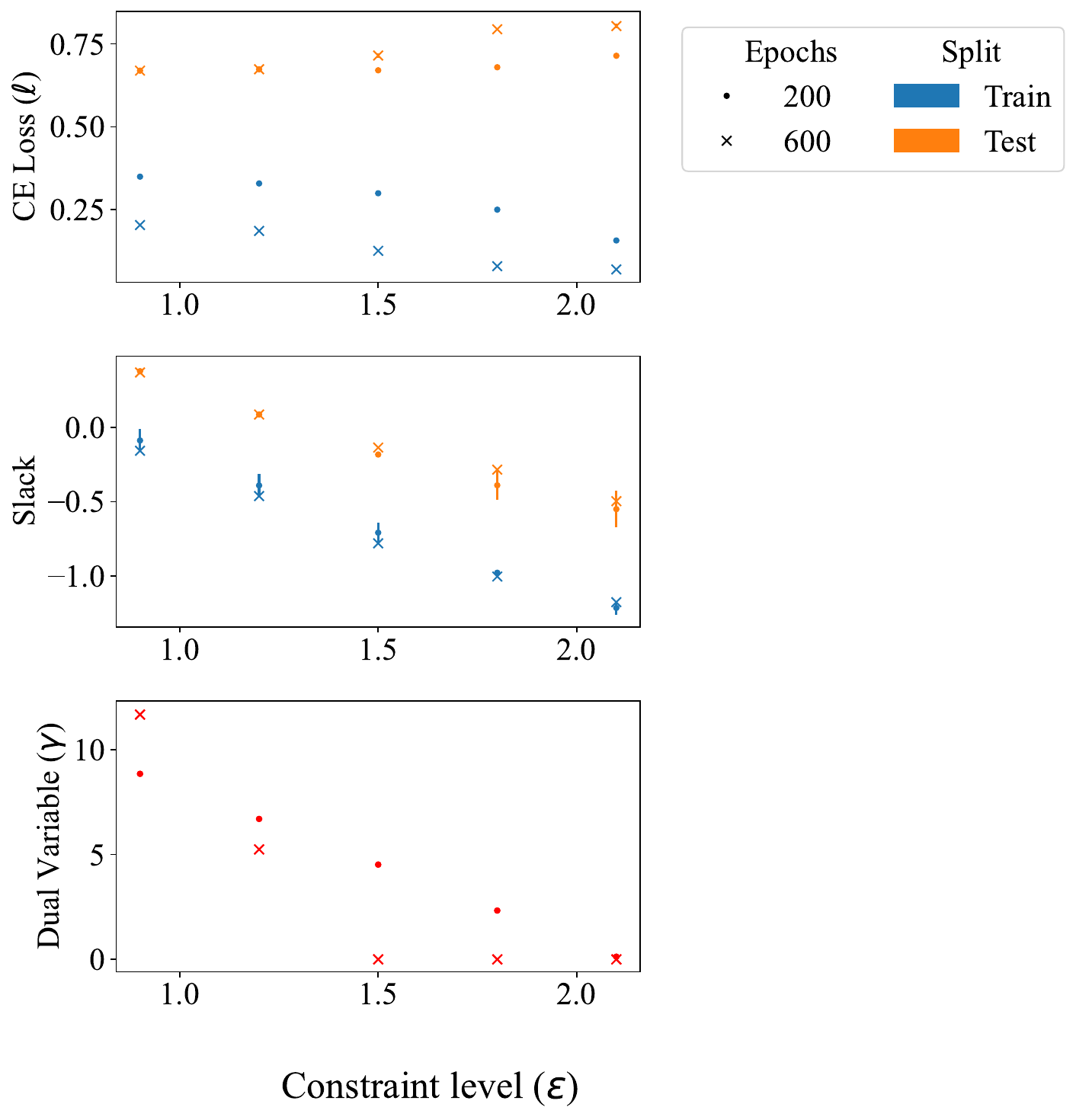}
        \caption{Training WideResnet-40-2 for 200 (standard) and 600  epochs in the CIFAR100 dataset, with different constraint levels.}\label{plot:daug:epochs}
\end{figure}

Figure~\ref{plot:daug:epochs} shows an ablation on the number of epochs. Non-zero dual variables and strict feasiblility show the algorithm has not yet converged at 200 epochs. At 600 epochs whereas constraint satisfaction shows little change,  training loss decreases and the generalization gap increases. Thus, we observe early stopping has a larger impact on the primal objective than on the constraint.
that although for stricter levels of the constraint the algorithm has not converged when training is stopped at 200 epochs, it can yield solutions that are still feasible and have a smaller generalization gap. 

\subsubsection{Sampling Steps Ablation}\label{a:sampling-steps}
\begin{figure*}[htpb]
\centering
     \includegraphics[width=.7\textwidth]{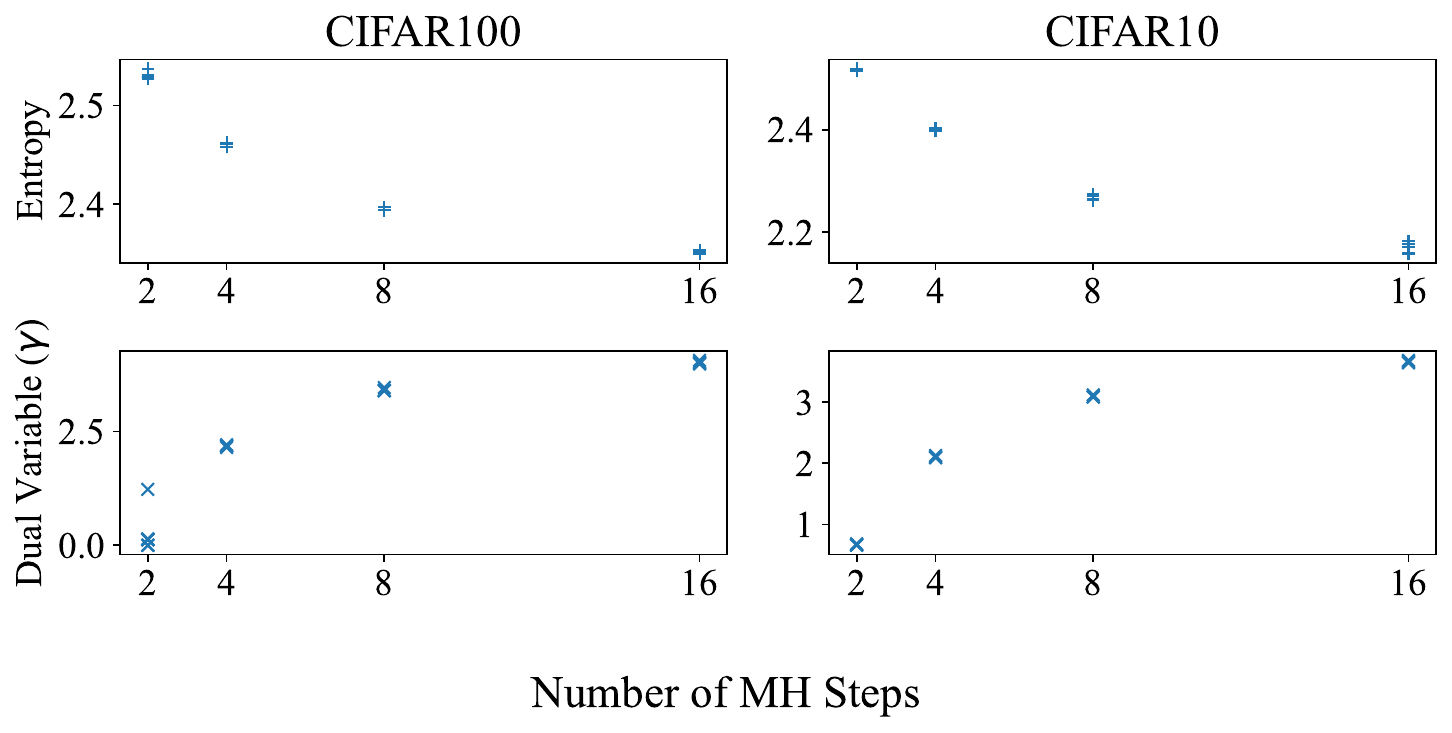}
        \caption{Number of Metropolis Hastings steps ablation for WideResnet-40-2 in CIFAR datasets. The constraint level is fixed for each dataset (0.8 in CIFAR10 and 2.1 for CIFAR100). We compute the entropy of the augmentations sampled at the last epoch of training. The value of the dual variable at epoch 200 increases with the number of steps, whereas the augmentation distribution entropy decreases. Each point represents an independent run. Experiments were repeated four times.}\label{a:plot:daug:steps-ablation}
\end{figure*}

\begin{figure}[htpb]
\centering
     \includegraphics[width=.45\textwidth]{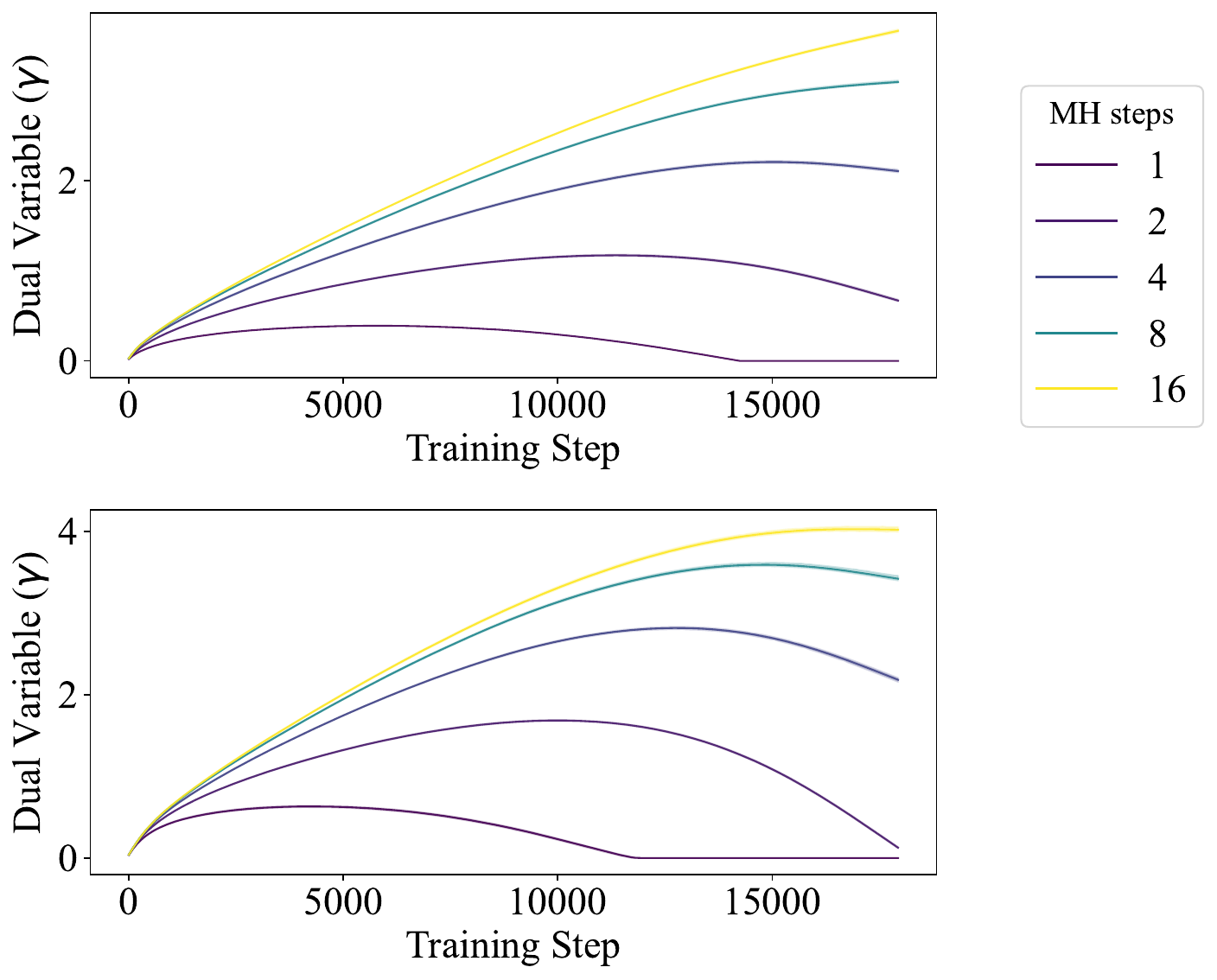}
        \caption{Evolution of dual variables during training for Wide-ResNet-40-2 in CIFAR datasets. The constraint level is fixed for each dataset (0.8 in CIFAR10 and 2.1 for CIFAR100). As the number of MH sampler steps increases so does the growth of dual variables, which reflects harder to satisfy  constraints.}\label{a:plot:dual-steps}
\end{figure}

As already mentioned in Section~\ref{proposed:algorithm}, we used the Metropolis-Hastings algorithm with independent uniform proposals. As shown in Figure~\ref{a:plot:daug:steps-ablation}, using more steps of the chain allows samples to deviate further from the uniform distribution, as measured by the decrease in entropy. As already mentioned, dual variables give information of the trade-off between fitting clean and augmented data. We observe the final value of the dual variable is highly correlated with the entropy of sampled transformations. That is, transformations that deviate further from uniformity result in larger dual variables at the end of training. Furthermore, in~\ref{a:plot:dual-steps} we show the evolution of dual variables for different sampling steps. Sampling distributions that are closer to worst case perturbations results in more stringent requirements, and thus dual variables grow more rapidly.

\begin{figure}[htpb]
\centering
     \includegraphics[width=.49\textwidth]{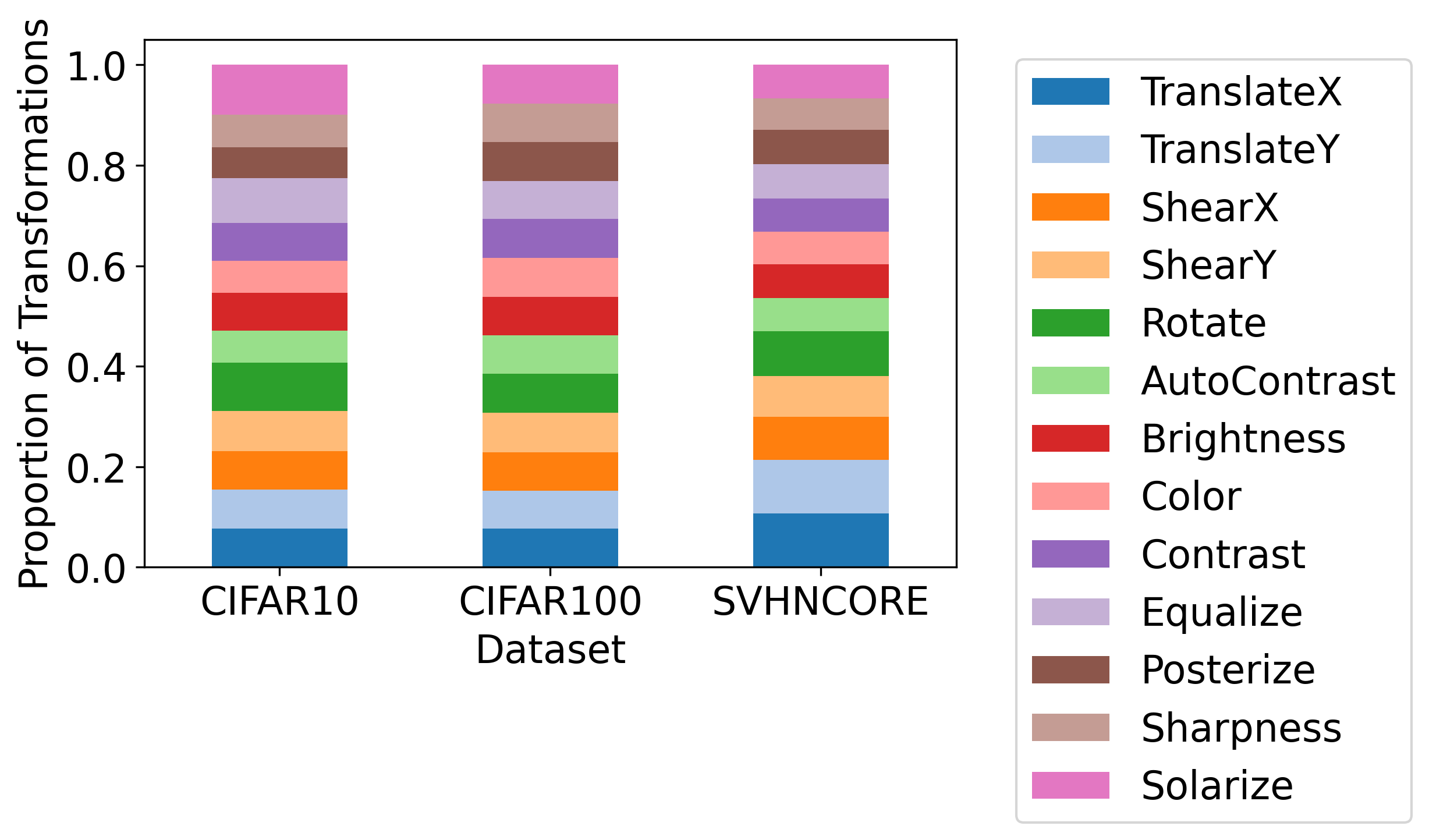}
        \caption{Frequency of sampled transformations for CIFAR and SVHN datasets using two MH sampler steps, for Wide-Resnet-28-10.}\label{a:plot:proportion}
\end{figure}
\subsubsection{Sampled Transformations}\label{sampled-transformations}
We now include plots of the empirical transformation distributions. As shown in Figure~\ref{a:plot:proportion}, the frequency with which transformations are sampled varies depending on the dataset, which is a desirable property. Similar to ~\cite{adversarial-auto-augment}, we observe a prevalence of geometric transformations, unlike~\cite{Auto-augment}. In SVHN color transformations are less frequent than in CIFAR datasets, and the frequency of geometric transformations increases, which is interesting due to the perceptual importance of shape in the digit recognition task.

\begin{figure*}[htpb]
\centering
     \includegraphics[width=.9\textwidth]{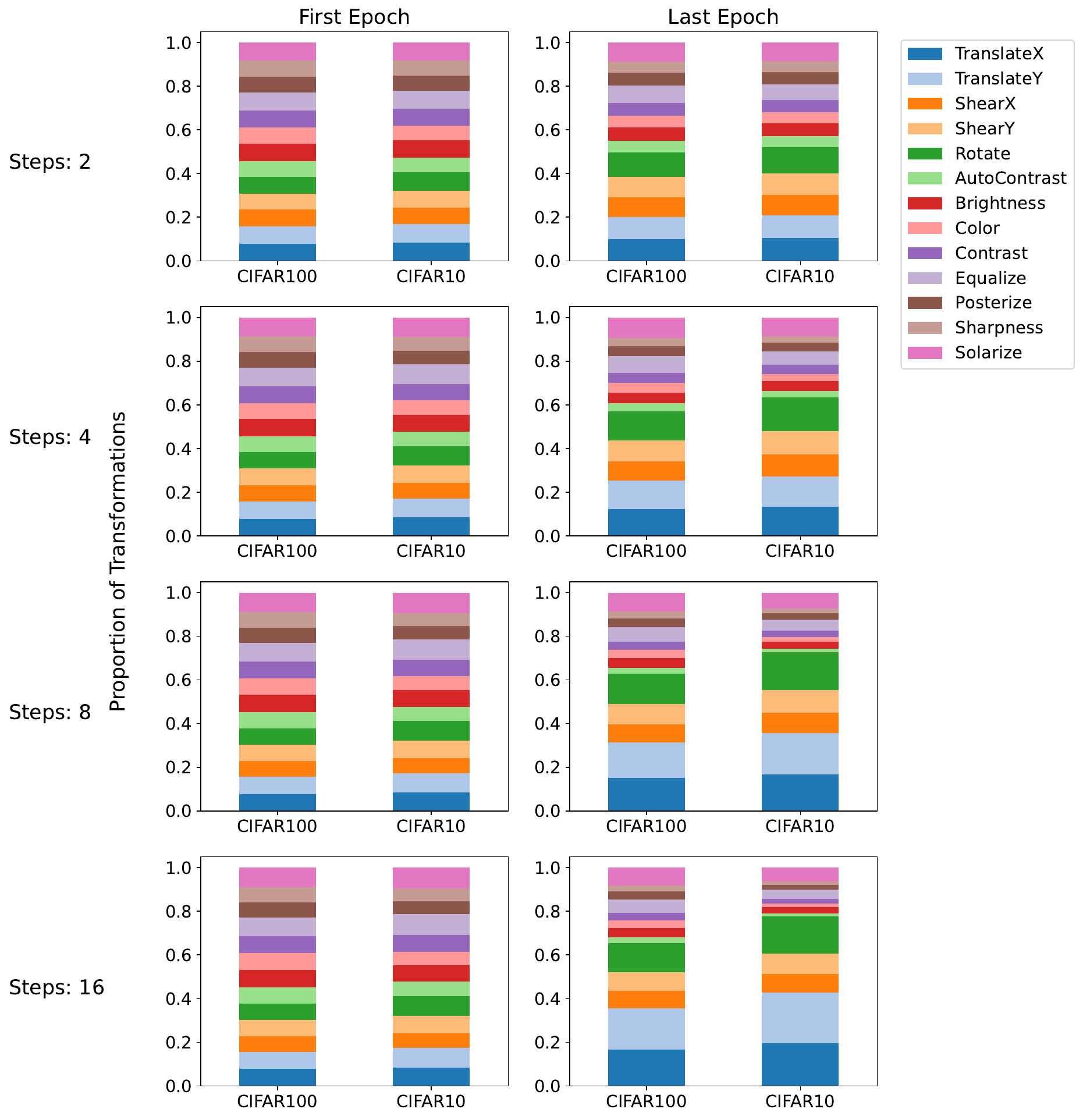}
        \caption{Frequency of sampled transformations for CIFAR datasets in the first and last epochs of training, for Wide-Resnet-40-2. As the number of steps increases, the the entropy of sampled transformations decreases, i.e., observed frequencies get further from uniformity. }\label{a:plot:proportion2}
\end{figure*}

As already noted, using more steps allows the chain to deviate further from the uniform proposal distribution. We thus plot the frequency of sampled  transformations across training epochs on Figure~\ref{a:plot:proportion2}. We observe that as training progresses, the observed frequencies also deviate further from uniformity. This suggests that, due to the dataset and architecture, some transformations may be harder to fit than others.

\begin{figure*}[htpb]
\centering
     \includegraphics[width=.95\textwidth]{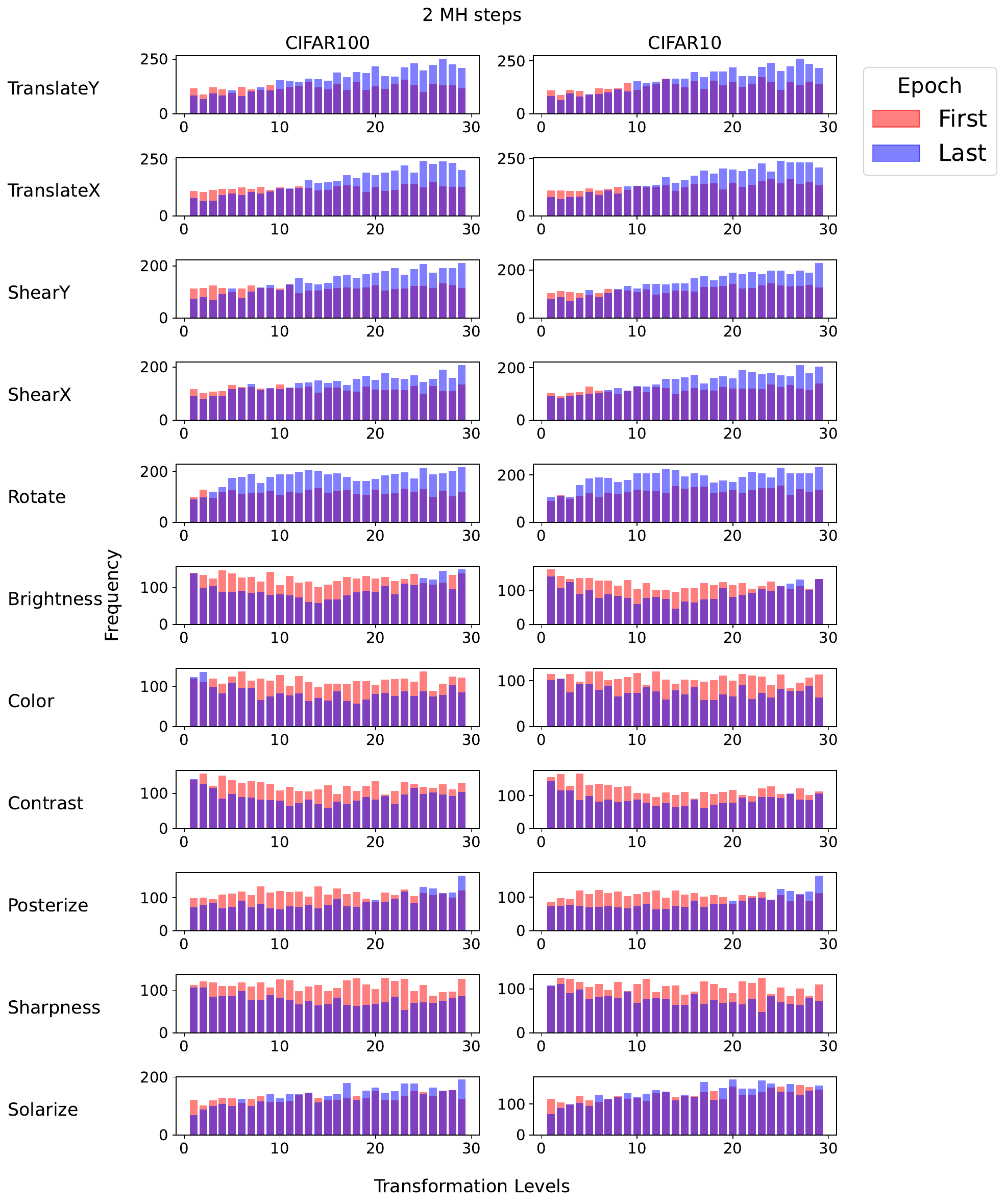}
        \caption{Frequency of sampled transformation levels across epochs, for different transformations, using two MH steps. Extreme levels are sampled more frequently, but some transformations deviate further from uniformity.}\label{a:plot:levels-2}
\end{figure*}

\begin{figure*}[htpb]
\centering
     \includegraphics[width=.95\textwidth]{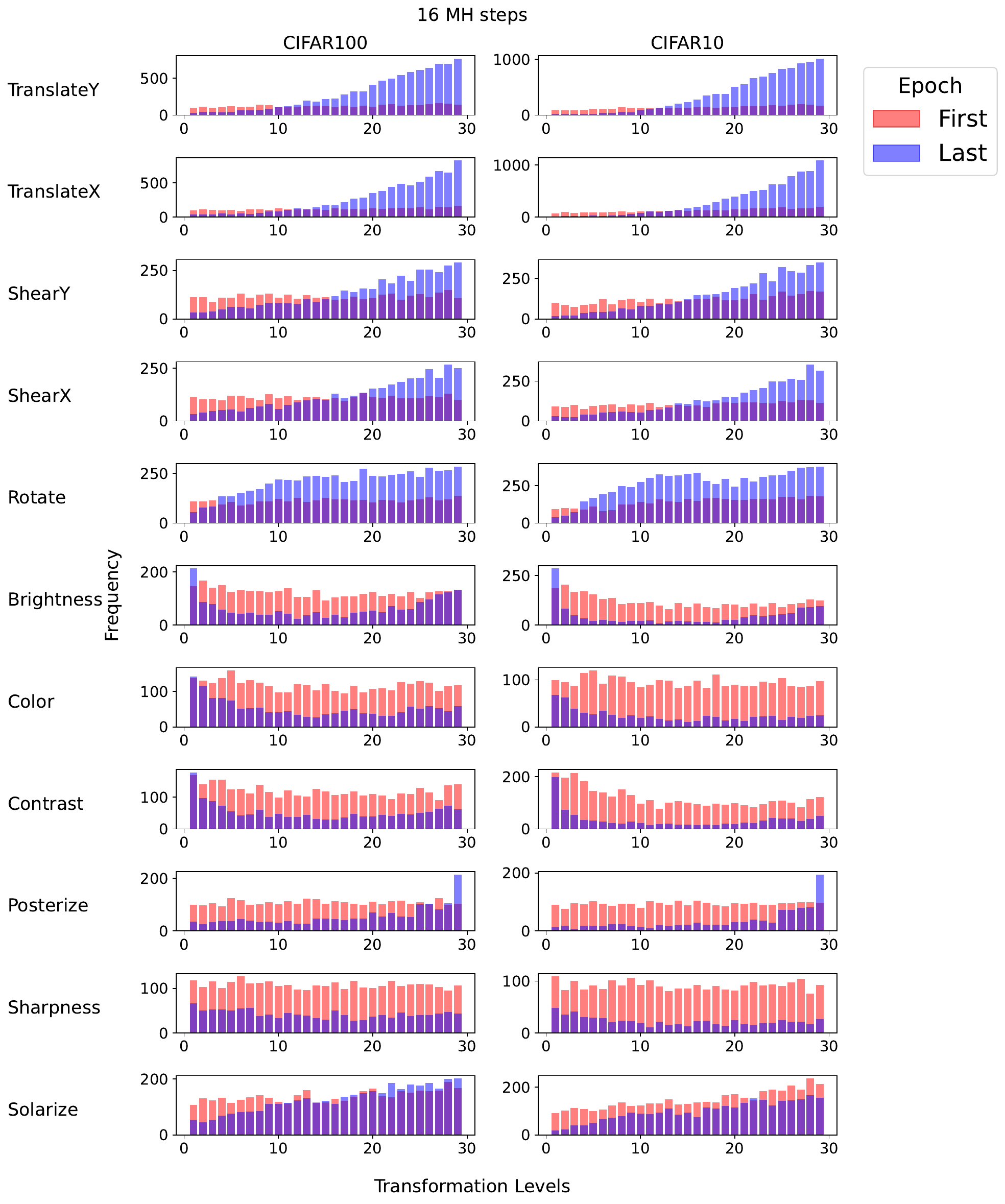}
        \caption{Frequency of sampled transformation levels for the first and last epochs, for different transformations, using sixteen MH steps. The frequencies concentrate in extreme values for some transformations.}\label{a:plot:levels-16}
\end{figure*}

Similarly, we include histograms for the sampled transformation levels in Figure~\ref{a:plot:levels-2}. Extreme levels are sampled more frequently, but the empirical distributions vary depending on the transformation. The deviation from uniformity and the differences between transformations are accentuated as sampling steps increase (Figure~\ref{a:plot:levels-16}).

\subsection{Runtime analysis}~\label{a:runtime}
As mentioned in section \ref{proposed:algorithm}, the added computational cost of our algorithm is that of computing a forward pass for each MCMC step. As a result, trade-off between sampling $\lambda^{\star}_c$ accurately and computation arises. In Table~\ref{table:runtime} we provide an empirical runtime analysis for our method for different numbers of MH steps, and compare it with the training time of baseline methods. These times correspond solely to training, and it is difficult to account for the time taken to tune the hyperparameters of each method, which hinders direct comparisons.  In the case of DeepAA~\citep{deep-aa}, it requires running a data augmentation policy search that takes 11 hours (more than $ 5 \times$ training time) using the same hardware. 

\begin{table}[htpb]
\centering
\setlength{\tabcolsep}{6pt} 
\renewcommand{\arraystretch}{1.3}
\begin{tabular}{c|cc|cc}
\hline
 & TA & DeepAA &\multicolumn{2}{c}{Ours}  \\
&  &  & 2 Steps & 4 Steps\\ \hline
Epoch time (s) & 12.6 & 13.3 &32.5 & 51.2 
\end{tabular}
\caption{Time per epoch for WideResnet 40-2 in CIFAR 10 dataset, on a workstation with one NVIDIA RTX 3090 GPU and AMD Threadripper 3960X (24 cores, 3.80 GHz) CPU. }\label{table:runtime}
\end{table}

\subsection{Batch mode Augmentation}\label{a:batch-mode}
All of the experiments were conducted using a single augmentation per training sample in the batch, that is $m=1$ in Algorithm~\ref{algo-mh}. As already mentioned, our approach is also suitable for batch mode augmentation, i.e. $m>1$. Therefore, conducted a preliminary ablation study on $m$. As shown in table~\ref{a:table:m-ablation}, increasing $m$ resulted in a slight decrease in accuracy. However, increasing $m$ leads to larger batch sizes, which affects optimization dynamics. Consequently, adjusting the learning rate and other optimization hyperparameters for each value of $m$ could lead to performance improvements, as shown by~\cite{deep-aa}. Therefore, providing further ablations and analyses regarding batch mode sampling is an interesting future research direction.
\begin{table}[]
\centering
\begin{tabular}{@{}ll@{}}
\toprule
m  & Accuracy \\ \midrule
2  & 81.58    \\
4  & 81.34    \\
8  & 81.25    \\
16 & 80.33    \\ \bottomrule
\end{tabular}
\caption{Test accuracy (second column) for WideResnet 40-2 trained on CIFAR100 using different values of augmentations per sample (denoted as $m$, first column).}\label{a:table:m-ablation}
\end{table}

\subsection{The bias of data augmentation can hinder performance}\label{a:data-aug-biases}
As already mentioned in section~\ref{rel-work-adv-aug} there is empirical evidence that certain distributions over commonly used transformations can introduce biases that are detrimental to model performance and generalization~\citep{trua-adversarial-aug}. We provide another simple experiment to show that in practice there exist transformation distributions over commonly used augmentation spaces that deteriorate model performance. We also showcase that balancing the amount of data augmentation (e.g. by including the original data) is important to mitigate and overcome this issue.

We restrict the transformations in the wide augmentation space~\citep{trivial-augment} to their maximum magnitude. We sample transformations according to $\lambda^{\star}_c$ using two MH steps as previously described. We compare against training without augmentation, and training using both augmentation and the original data (i.e. adding the identity) equally weighted. While sampling transformations according to $\lambda^{\star}_c$ deteriorates performance with respect to the model without augmentation, including both the identity and the augmented samples achieves a superior performance, as shown in Table~\ref{table:augmentation-is-bad}.
\begin{table}[]
\centering
\setlength{\tabcolsep}{4.5pt} 
\renewcommand{\arraystretch}{1.3}
\scalebox{1.0}{
\begin{tabular}{lll}
No augmentation & $\lambda^{\star}_c$ & $\lambda^{\star}_c$ + Identity \\ \hline
$78.42\pm0.31$  & $75.19 \pm0.54 $ & $80.01\pm0.26$            
\end{tabular}}
\caption{Image Classification test accuracy for WideResnet 40-2 in CIFAR100, trained using different augmentation policies defined over the wide~\citep{trivial-augment} augmentation space. The first column corresponds to using the training data without applying any transformations, and the second column to sampling transformations according to $\lambda^{\star}_c$, which results in lower accuracy. The third column corresponds to using both the original and augmented data equally weighted, which leads to an improvement in accuracy. We report the mean and confidence intervals computed over five independent runs.}\label{table:augmentation-is-bad}
\end{table}

\newpage
\subsection{Constraint Level Specification.}~\label{a:epsilon}

\begin{table*}[hpb!]
\centering
\begin{tabular}{@{}lllll@{}}
\toprule
Dataset & Architecture & Invariance     & Same $\epsilon$  & Adjusted $\epsilon$ \\ \midrule
MNIST   & MLP          & Translation    & $93.5 \pm 0.2$   & $94.9 \pm 0.1$      \\
        &              & Rotation (90)  & $96.95 \pm 0.03$ & $97.28 \pm 0.07$    \\
        &              & Rotation (180) & $93.18 \pm 0.03$ & $95.35 \pm 0.03$    \\
        & CNN          & Translation    & $96.00 \pm 0.02$ & $97.79 \pm 0.09$    \\
        &              & Rotation (90)  & $98.15 \pm 0.09$ & $98.5 \pm 0.1$      \\
        &              & Rotation (180) & $96.3 \pm 0.3$   & $97.2 \pm 0.3$      \\
F-MNIST & MLP          & Translation    & $76.5 \pm 0.5$   & $81.2 \pm 0.2$      \\
        &              & Rotation (90)  & $81.8 \pm 0.1$   & $84.78 \pm 0.06$    \\
        &              & Rotation (180) & $76.1 \pm 0.7$   & $83.85 \pm 0.21$    \\ \bottomrule
\end{tabular}
\caption{Classification accuracy for synthetically invariant datasets using the same constraint level for all transformations (third column) and adjusting constraint levels according to dual variables (fourth column).}~\label{a:table-epsilon-heuristic}
\end{table*}
This experiment illustrates how dual variables can be used to adjust epsilon using simple heuristics. In the synthetic invariances setup, we first train the model with the same constraint level for all transformations, regardless of whether the transformation is a ``true'' invariance of the data distribution. That is, there is a mismatch between the constraints and the invariances in the datasets, which can result in poor performance. We then proceed to: 
\begin{enumerate}
    \item[(i)] Tighten the constraints associated with zero dual variables by dividing epsilon by a factor of two.
    \item[(ii)] Relax the remaining constraints, which showed large associated dual variables, by multiplying $\epsilon$ by a factor of two.
\end{enumerate}
As shown in Table~\ref{a:table-epsilon-heuristic}, this simple heuristic leads to an improvement in test accuracy over the initial (misspecified) constraint levels. Therefore, adapting the constraint level by leveraging dual dynamics while training is a promising future work direction.

\FloatBarrier


\end{document}